\documentclass{article}

\usepackage{tabularx}
\usepackage{hyperref}
\usepackage{booktabs}
\usepackage{multirow}
\usepackage{tikz}
\usepackage{pgfplots}
\usepackage{standalone}
\usepackage{xcolor}
\usepackage{standalone}
\usepackage{longtable}
\usepackage{caption}
\usepackage{subcaption}
\usepackage{float}
\definecolor{darkblue}{rgb}{0, 0, 0.5}
\hypersetup{colorlinks=true,citecolor=darkblue, linkcolor=darkblue, urlcolor=darkblue}
\usepackage[belowskip=-5pt,aboveskip=0pt]{caption}
\usepackage[margin=1in]{geometry}
\usepackage{natbib}
 \setcitestyle{aysep={}}	% no comma between author and year
 \renewcommand{\cite}{\citep}	% to get "(Author Year)" with natbib
 \newcommand{\namecite}{\citet}	% to get "Author (Year)" with natbib

\makeatletter
\def\@maketitle{%
  \newpage
  \null
  \vskip 0em%
  \begin{center}%
  \let \footnote \thanks
    {\LARGE \@title \par}%
    \vskip 1.5em%
    {\large
      \lineskip .5em%
      \begin{tabular}[t]{c}%
        \@author
      \end{tabular}\par}%
    \vskip 1em%
  \end{center}%
  \par
  \vskip 1.5em}
\makeatother
\usepackage{etoolbox}
\makeatletter
\patchcmd{\@maketitle}{\@date}{}{}{}
\patchcmd{\@maketitle}{\vskip 1.5em}{\vspace{-2em}}{}{}
\makeatother
\usepackage{academicons}
\definecolor{orcidlogocol}{HTML}{A6CE39}
\newcommand{\appendixsection}[1]{\addtocounter{section}{1}%
   \setcounter{table}{0}
   \setcounter{figure}{0}
   \setcounter{equation}{0}
  \section*{Appendix \Alph{section}: #1}%
}
\renewcommand\appendix{%
   \setcounter{section}{0}
   \renewcommand{\theequation}{\Alph{section}.\arabic{equation}}
}
\pgfplotsset{compat=1.15}
\begin{document}

\title{\vspace{0cm} \bf \Large \rule{\textwidth}{3pt}
\begin{center}
{\bf \Large A data-centric approach to class-specific bias \\ in image data augmentation}
\end{center}
\rule{\textwidth}{2pt}
}

\author{
\begin{tabular}{cc} \\ \\
Athanasios Angelakis & Andrey Rass \\
Amsterdam University Medical Center & Den Haag, Netherlands \\
University of Amsterdam - Data Science Center & anvrass@gmail.com \\
Amsterdam Public Health Research Institute &  \\
Amsterdam, Netherlands &  \\
a.angelakis@amsterdamumc.nl & \\
\end{tabular}
}
\date{} % Empty date if you don't want it to show
\maketitle

%\begin{center}
%\large{\yourname{}} \\
%\end{center}

\begin{abstract}
Data augmentation (DA) enhances model generalization in computer vision but may introduce biases, impacting class accuracy unevenly. Our study extends this inquiry, examining DA's class-specific bias across various datasets, including those distinct from ImageNet, through random cropping. We evaluated this phenomenon with ResNet50, EfficientNetV2S, and SWIN ViT, discovering that while residual models showed similar bias effects, Vision Transformers exhibited greater robustness or altered dynamics. This suggests a nuanced approach to model selection, emphasizing bias mitigation. We also refined a "data augmentation robustness scouting" method to manage DA-induced biases more efficiently, reducing computational demands significantly (training 112 models instead of 1860; a reduction of factor 16.2) while still capturing essential bias trends.
\end{abstract}

% EXCLUDED: Moreover, the reduced granularity of our experiments still highlighted the relevant trends, answering the original paper's question of whether the approach's heavy computational requirements could be lessened.

\section{Introduction}
% EXCLUDED: Machine learning and data science have become increasingly prevalent in multiple industries in recent years, thanks to advancements in both technology and theory, as well as the availability of vast amounts of data. In particular, computer vision, being a common application, has proven to be a powerful tool to address a variety of sub-tasks, such as object detection and image classification, fuelling innovation in a variety of fields - from medical diagnostics to self-driving cars. %filler

Machine learning is generally defined as aiming at systems to solve, or "learn" a particular task or set of tasks (e.g. regression, classification, machine translation, anomaly detection - \cite{Goodfellow-et-al-2016a} based on some training data, allowing computers to assign labels or predict future outcomes without manual programming. 
A typical case involves a training dataset that is finite, and the system’s parameters are optimized using a technique such as gradient descent, using some performance measure (e.g. “accuracy”) for evaluation during optimization and on a holdout “test” set \cite{lecun1998gradient, bishop2006pattern, shalev2014understanding, Goodfellow-et-al-2016a}.  % This entire paragraph is paraphrased from Balestriero. Not sure if we should leave it as is, or remove altogether, or rewrite heavily.

In particular, computer vision tasks (e.g. image classification) are contemporarily accomplished via deep learning methods \cite{feng2019computer} such as
% EXCLUDED: Moreover, the reduced granularity of our experiments still highlighted the relevant trends, answering the original paper's question of whether the approach's heavy computational requirements could be lessened.
% Deep learning is a broad subfamily of machine learning methods based on approximating the desired "learned" function by composing together multiple functions or "layers" into artificial "neural networks". The advantage of such an approach is the ability to approximate highly complex functions with this method, given the correct number and configuration of functions \cite{lecun2015deep}. The "learning" then, simply, lies in optimizing the parameters of the network (typically, via some form of gradient descent) to achieve the best approximation of the target function \cite{Goodfellow-et-al-2016b}. In particular, 
convolutional neural networks (CNNs), which have continuously been held in high regard as the go-to approach to such problems \cite{lecun1995convolutional, lecun2010convolutional, voulodimos2018deep}, owing to their ability to extract features from data with known grid-like topology, such as images \cite{Goodfellow-et-al-2016c}.

Like any other machine learning systems, CNNs can suffer from overfitting - a performance gap between training and test data samples, which represents inability to generalize the learned task to unseen data \cite{Goodfellow-et-al-2016b}. To combat this, various regularizaton methods have been developed for use during optimization \cite{tikhonov1943stability, tihonov1963solution}.  Particular (though not limited - \cite{ko2015audio}) to image-based tasks is the use of data augmentation - applying certain transformations, such as random cropping, stretching and color jitter, to training data during training iterations. Such techniques are near-ubiquitous in computer vision tasks due to their effectiveness as a regularization measure \cite{shorten2019survey}.

However, recent research by \namecite{NEURIPS2022_f73c0453} suggests that, despite data augmentation being such a prevalent method of improving model performance, it may actually prove a risk to blindly turn to this technique regardless of dataset and approach. This appears to be a part of a larger phenomenon concerning a tendency of parametrized regularization measures to sacrifice performance on certain classes in favor of overall model accuracy. However, it is also caused in large part by the fact that different image transformations seem to possess varying levels of label preservation \cite{10.1109/TASLP.2015.2438544, taylor2018improving} dependent on the class of image data, and, as such, can have a severe class-specific negative impact on the model performance by virtue of incurring label loss if applied too aggressively. This impact can also be so deep as to even impact downstream performance in the case of transfer learning tasks.

It may be a natural reaction, as such, to caution against using data augmentation as a regularization technique to avoid this. However, this can prove to be difficult in practice, as the overall performance boost this technique provides is undeniable, and it is currently used ubiquitously, with few alternatives. In addition, as alluded to before, the paper   showed that other regularization methods, such as weight decay, can also be subject to this phenomenon. It is also important to note that \namecite{NEURIPS2022_f73c0453} claims that the phenomenon is model-agnostic for popular CNNs based on residual blocks such as ResNet \cite{he2015deep}, DenseNet \cite{DBLP:journals/corr/HuangLW16a}, and others, % can shorten this example list
but does not make any claims in regards to whether the phenomenon is data-agnostic, or how it would manifest in image classification networks that belong to architectures founded in considerably different principles, such as Vision Transformers \cite{dosovitskiy2020image, Liu_2021_ICCV} that use patch-based image processing and self-attention mechanisms to extract features from images.

Our work follows a course of further investigation. To formulate our primary research question, we follow the tenets of the data-centric AI movement championed by Andrew Ng, which seeks to systematically engineer the data used in training AI systems, and places special focus on accounting for imperfections in real world data. In data augmentation, the discipline outlines issues such as domain gaps, data bias and noise. Following this school of thought leads us to question if class-specific bias from data augmentations affect datasets different from Imagenet, in a different way. In particular, we seek to test whether this phenomenon can be observed on datasets that differ in nature to various degrees from Imagenet \cite{deng2009imagenet}, which was used in Balestriero’s 2022 paper, and to what extent. To supplement this line of validation, another, secondary research question emerged from a seemingly minor detail in the original study. Concretely, we investigate if the addition of Random Horizontal Flipping have an effect on how the class-specific bias phenomenon manifests.

While not our core focus, we also seek to confirm how model-agnostic this phenomenon is on these new datasets. For this, it is first necessary to test with a model that has common features with ResNet50, which was the baseline for many experiments in \cite{NEURIPS2022_f73c0453}. We selected EfficientNet, first described in \namecite{pmlr-v97-tan19a}, which is a family of models that also utilizes residual blocks, but was designed via a neural architecture search \cite{elsken2019neural} using a new scaling method that uniformly scales all dimensions of model depth/width/resolution using a simple yet highly effective compound coefficient. To that end, we question if class-specific bias from data augmentations on the same dataset would affect a different Residual CNN architecture in the same manner as it would a ResNet. Finally, it is also worth investigating the effects of a vastly different architecture, as mentioned earlier. For these purposes, we have selected to use a SWIN Transformer, which is a relatively small patch-based vision transformer that uses a novel shifted windowing technique for more efficient computation of the self-attention mechanism that is inherent to Transformer-type models \cite{Liu_2021_ICCV}. Finally, we consider if class-specific bias from data augmentations on the same dataset would affect a Vision Transformer model in the same manner as it would a ResNet. % It may be worth condensing these two research questions into one, as we will only have one chapter for the both of them.
% I tried to make the introduction shorter (I did take it down to 1 page) but making it as short as Balestriero is difficult, unless I throw away some of the context or base-level explanations such as "what is machine learning"?
% For now I will leave this as-is, but I will cut out the basic pointlessness if we don't have space

%%%%%%%%%%%%%%%%%%%%%%%%%%%%%%%%%%%%%%%%%%%%%%
%%%%%%%%%%%%%%%%%%%%%%%%%%%%%%%%%%%%%%%%%%%%%%
\section{The Effect Of Data Augmentation-Induced Class-Specific Bias Is Influenced By Data, Regularization and Architecture}

This section details our study's data-centric and model-centric analysis of the phenomena originally observed in \cite{NEURIPS2022_f73c0453}. Firstly, we establish a practical framework for replicating such experiments in Section 2.1. Following this, we use a ResNet50 model trained from scratch with the Random Cropping and Random Horizontal Flip DA to provide the data-centric analysis of DA-induced class-specific bias on three datasets (Fashion-MNIST, CIFAR-10 and CIFAR-100) in Section 2.2. We then take a step back in Section 2.3 to evaluate the potential side effects of including the Random Horizontal Flip augmentation, as done in the original study. Finally, we conclude by demonstrating how alternate computer vision architectures interact with the phenomenon illustrated in previous sections. These findings are key as they serve to deepen our understanding of the potential pitfalls of introducing DA to computer vision tasks in order to improve overall model performance, while showing how the problem of class-specific bias can be alleviated or forestalled.

%%%%%%%%%%%%%%%%%%%%%%%%%%%%%%%%%%%%%%%%%%%%%%
\subsection{Data Augmentation Robustness Scouting}
% If we want we should outline here a framework based on our methodology (instead of describing our methodology in gruesome detail) so that our conclusion makes sense and so that we make another contribution with this (see conclusion comments)
In this section, we seek  to formalize a "bare minimum" procedure necessary to adopt and replicate an experiment framework for assessing the tradeoff between overall model performance, DA intensity and class-specific bias by outlining the specifics of our experiments' practical implementation. The goal of this is to serve as a guideline for applying the findings of \namecite{NEURIPS2022_f73c0453} in a more efficient manner that is better fit for practical or "business" environments, as well to lay the groundwork for procuring the results that shall be discussed in further sections of this chapter.

We propose the following procedure, further referred to as "Data Augmentation Robustness Scouting": First, a set of computer vision architectures is to be selected for a given dataset and DA regimen. Following this, the model is trained on a subset of the data in several training runs, such that each run features an increasing intensity of augmentation (represented by as a function of $\alpha$). The test set performance (per-class and overall) is then measured for every value of $\alpha$, such that dynamics in performance can be observed as $\alpha$ is gradually increased. This procedure is then repeated from start to end, with the performance being averaged out for every value of $\alpha$ to smooth out any fluctuations resulting from the stochastic nature of the training process. The granularity (expressed through the amount and range of $\alpha$ steps and amount of runs per value of $\alpha$) should be kept to a bare necessary minimum such that the desired clarity required to observe trends in performance dynamics over different $\alpha$ values can be achieved. Following this study, if "ranges of interest" in $\alpha$ are established, it is recommended to perform the described procedure again with higher granularity within those ranges.

The experiments detailed in further sections of this chapter featured the DA Robustness Scouting procedure performed on three datasets - Fashion-MNIST, CIFAR-10 and CIFAR-100 \cite{xiao2017fashion, krizhevsky2009learning}. The Random Cropping DA was evaluated, along with Random Horizontal Flip as a fixed supplementary augmentation. For our purposes, the preliminary model tuning and experiments were implemented using the Tensorflow and Keras libraries for Python. Our approach was based on best optimization practices for training convolutional neural networks as described in \namecite{Goodfellow-et-al-20167}. Before an experiment can be run, a given model is be repeatedly trained from scratch on a given dataset with a set of different permutations of learning rate, maximum epoch and batch size parameters. This was done in order to achieve a “best case” base model - one that did not use any overt regularization (so as not to obscure the impact of regularization during the experiment), provided the best possible performance on the test set, and was the least overfitted. As batch size tuning can be considered a form of regularization, only minimal tuning was applied to batch size to ensure sufficient model stability and performance. A validation subset (10\% of the training set - common value used) was used to evaluate overfitting. It is important to note that the definition of “sufficient” is very dataset-specific, as some datasets lend themselves much more easily to being “solved” by models such as ResNet50 - both in terms of training error and generalization. Typically, regularization is used for the very purpose of helping improve generalization, but in the case of this study doing so would risk obscuring the phenomenon being observed. As such, a decision was made to “settle” that “best” base performance on some datasets would be considered subpar ceteris paribus.

The ResNet50 architecture was directly downloaded from Tensorflow’s computer vision model zoo sans the Imagenet weights, as the models would be trained from scratch.  \namecite{NEURIPS2022_f73c0453} makes no mention of the optimizer used, so the Adam optimizer \cite{kingma2014adam} was used in all tuning and experiment steps. Each combination of model and dataset was tuned across a set of learning rates, epoch counts and batch sizes selected with the goal of getting the best validation and test set accuracies (using the sparse\_categorical\_accuracy metric from Keras) while being mindful of models' tendency towards overfitting.

The "small" EfficientNetV2S \cite{DBLP:journals/corr/abs-2104-00298} architecture was selected for its respective experiment due to being a modern and time-efficient implementation of the EfficientNet family of models. The model was downloaded, tuned and optimized as per the above procedure. 

The SWIN Transformer architecture was implemented based on existing Keras documentation, which was inspired by the original approach used in \namecite{Liu_2021_ICCV}, processing image data using 2x2 patches (about 1/16 of the image per patch, similar to \namecite{DBLP:journals/corr/abs-2010-11929}) and using an attention window size of 2 and a shifting window size of 1. However, this approach originally included built-in regularization methods that deviated from our base experiment structure, such as random cropping and flipping already built into the model, label smoothing and an AdamW optimizer \cite{loshchilov2017decoupled}, which features decoupled weight decay. These were all omitted from the base model used in tuning and further experiments, while maintaining the core architecture.

As it is generally widespread and commonly used as a data augmentation technique, the particular Random Crop procedure to be used was not well-defined in the text of \namecite{NEURIPS2022_f73c0453}. As such, for the purposes of this study, the Random Crop data augmentation was defined as applying the “random crop” transformation from Tensorflow’s image processing library to the training dataset at every iteration, with the resulting image height and width calculated using the following formula:
\[new\_image\_size = round(image\_size*(1-\alpha))\]
Where $\alpha$ is a percentage or fraction representing the portion of the image that would be omitted and round() is the default Python. No padding was used. After being cropped, the images were upscaled back up to their original size so that they would still match the input layer requirements of 32x32px that the model trained on them required. In addition, these dimensions also mean that the test images do not need to be cropped down to the new size, which would be the more likely scenario in practice.

Finally, in order to accommodate existing limitations in available computing, the granularity of the experiments was reduced from 20 models per Random Crop transformation alpha to 4, as well as adjusting $\alpha$ in steps of 3-4\%, rather than evaluating every 1\%. To ensure the results maintained integrity, the exact figures for this granularity reduction were motivated by finding the minimum number of runs which corresponded to insignificant marginal deviations in the resulting mean test accuracy per additional run, preserving the expected trend. The new granularity of the augmentation alpha was determined based on the observation that the original paper’s highlighted trends in accuracy would still persist if this smoothing was applied to them, if not become simpler to detect by virtue of eliminating the existing fluctuations. In addition, using steps of Random Crop $\alpha$ that were finer would produce the same dimensions in post-crop training images due to the rounding involved and the images’ small size, leading to multiple iterations which would be practically identical, thus creating redundancy. 

Additionally, in the case of all conducted experiments, the Keras callback implementation of Early Stopping was used with a 10\% validation subset separated from the main training subset. It is a popular deep learning technique that monitors some metric after each training epoch (typically, loss or a select accuracy metric on the validation subset) and terminates training if it fails to perform better than a previous evaluation in a number of epochs described by the “patience” parameter. This algorithm also commonly features an optional restoration of model weights to the best-performing epoch. Early Stopping is commonly recognized as a regularization technique - however, it is considered unobtrusive, as it requires almost no change in the underlying training procedure, and can also be thought of as a “very efficient hyperparameter selection algorithm” \cite{Goodfellow-et-al-20167}. As this work concerns the effects of regularization, a decision was made that Early Stopping must be used cautiously,  to minimum effect,  despite its unobtrusive nature. 
%EXCLUDED: As the amount of models trained per random crop alpha was limited, and the models experienced mild instability due to the size and dimensions of the training datasets that were used, the technique was applied to ensure the relative “ideal” model was achieved for each run. The motivation was not to stop training early, but rather to make use of the “restore best weights” functionality. As such, the “patience” parameter for the Early Stopping callback was selected individually for each experiment to minimize the occurrence of actual premature termination in the learning process by first tuning the model without the callback and then setting the patience to either exceed the amount of epochs remaining from the “best” epoch until the end of training, or to at least equal the amount of epochs typically necessary for the model to converge to such a point. In addition, there is a risk of minor sacrifices in generalization quality compared to runs where the model is allowed to train for considerably longer training times - a study conducted by \cite{prechelt2002early} concluded “that slower stopping criteria allow for small improvements in generalization (here: about 4\% on average), but cost much more training time (here: about factor 4 longer on average)”. Such a trade-off was deemed negligible considering the purpose and limitations of our paper. %Do we want to include this bit of context?

%%%%%%%%%%%%%%%%%%%%%%%%%%%%%%%%%%%%%%%%%%%%%%

\begin{figure}[ht]
\centering
\includegraphics[width=1\textwidth]{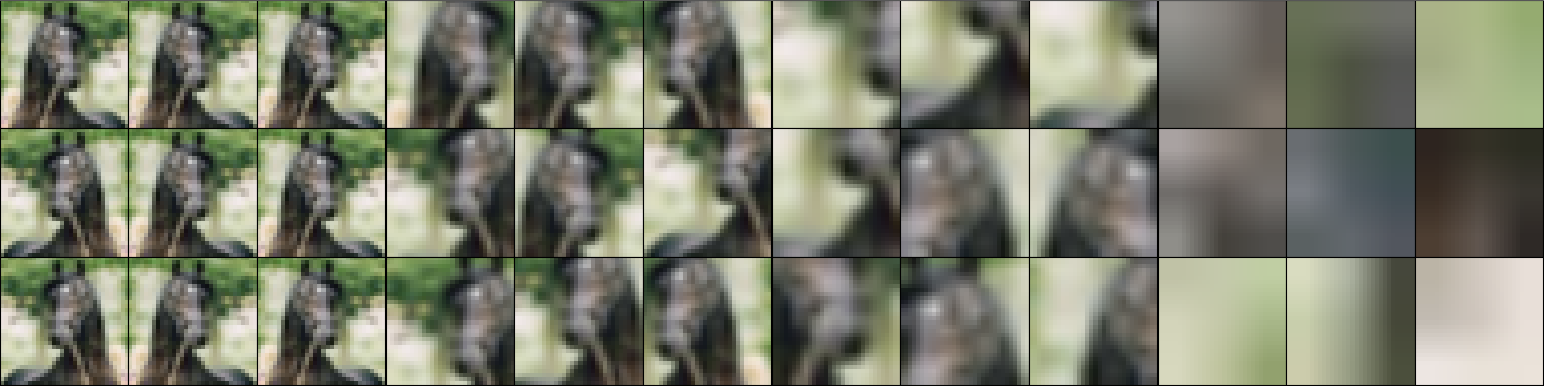}
\label{fig:0-30-60-90}
\caption{This figure shows a collection of four 3x3 grids, each demonstrating the effect of a Random Crop + Random Horizontal Flip combined data augmentation on an image belonging to the "horse" class of the CIFAR-10 dataset. The Random Crop $\alpha$ for is 0\%, 30\%, 60\% and 90\% for each grid, respectively. Label loss progression can be intuited to a degree, as the proportion of images that lack the distinctive features of the "horse" class increases with $\alpha$.
}
\end{figure} 
%%%%%%%%%%%%%%%%%%%%%%%%%%%%%%%%%%%%%%%%%%%%%%
\subsection{The Specifics Of Data Affect Augmentation-Induced Bias}

As a key part of our data-centric analysis of the bias-inducing effects of DA detailed by \namecite{NEURIPS2022_f73c0453}, we conducted a series of experiments based on the paper's initial proposal of training and evaluating a set of CNN models while adjusting a DA regime represented as a function of some parameter $\alpha$ between batches of runs (see Figure 2 for results). To focus the scope of our research, we have limited the experiments in this section to using ResNet50 as the architecture of choice, with Random Horizontal Flipping applied combined with Random Cropping with increasing portions of the training images obscured as the parametrized augmentation. For illustrative purposes, three datasets were chosen for the purposes of the trial to provide a relative diversity of content: Fashion-MNIST, CIFAR-10 and CIFAR-100. All three datasets differ from ImageNet in possessing a much smaller image quantity ($<100K$ vs. $>1M$), class count ($\leq100$ vs. $>1K$), and image size ($<40$px vs. $>200$px). There are some additional distinctions: Fashion-MNIST consists of greyscale top-down photos of clothing items, whereas CIFAR-100 has very few images per class, despite CIFAR datasets being otherwise quite close in subject matter to ImageNet.

Despite significantly fewer runs, the results of our experiments do clearly illustrate the label-erasing effect of excessive DA application and its strong variation between classes, as seen, for example, in the performances of the Coat, Dress, Shirt and Sandal in Figure 2, each of which experiences distinctive performance dynamics as $\alpha$ increases, and each category has a different threshold of $\alpha$ past which label-loss from Random Cropping occurs rapidly (as illustrated by a rapid drop-off in test set accuracy). In addition to confirming class-specific DA-induced bias, we observe a very clear difference across the three datasets in terms of the speed at which this effect is seen in mean performance, as well as the degree of difference between individual classes. This can be attributed to the complexity of each dataset, but also clearly illustrates how "robustness" to class-specific bias from DA can vary heavily between datasets in the same way as it can between classes. While this difference is noticeable between the three observed datasets, it is even more stark when compared to ImageNet's performance in \namecite{NEURIPS2022_f73c0453}, as label loss and corresponding mean test set performance decline occur at much earlier values of $\alpha$ in our observations. 

As in \namecite{NEURIPS2022_f73c0453}, mean test set performance in all cases follows a trend of "increase, fall, level off" as $\alpha$ pushes all but the most robust classes past the threshold of complete label loss (for example, in the Figure 2, the mean accuracy on CIFAR-10 reaches its highest point of 0.764 at 10\% $\alpha$, then rapidly declines until an $\alpha$ of 70\%, after which it levels off at around 0.21. On the other hand, the Fashion-MNIST plot is very illustrative of class-specific behaviors, as we see the "Sandal" class reach its peak accuracy of 0.994 at an $\alpha$ as late as 36\%, while "Coat" begins to drop from 0.86 down to near-zero accuracies as early as 10\% and 43\%, respectively. For a full summary of the diverse $\alpha$ at which each class and mean test set performance reach their peak, see Appendices F, G and H.

The causes of this dataset-specific bias robustness can be narrowed to two causes - overall complexity as a learning task (e.g. Fashion-MNIST is "simple to solve", and thus experiences minimal benefit from regularization before reaching detriment), as well as robustness to label loss from a given DA, which can occur on dataset or class level. Class-level robustness can be illustrated by comparing the performance of the T-Shirt and Trouser classes in Figure 2, as images belonging to the Trouser class are quite visually distinct from most other categories even at higher levels of cropping, while the T-Shirt class quickly loses its identity with increased $\alpha$ values, manifesting in the visible performance difference. Dataset-level robustness can occur due to such factors as training images containing more information (e.g. "zoomed-out" depictions of objects, higher image size, as well as RGB color), and can be seen in how the CIFAR datasets' mean performance and its dynamics relative to $\alpha$ compare to Fashion-MNIST. 

Here, a promising direction for research could be to conduct a range of similar experiments on datasets picked out specifically with this robustness to a particular DA in mind - for example, a dataset such as the Describable Textures Dataset \cite{cimpoi14describing} consists of images of objects with a focus on texture, with 47 classes, such as "braided", "dotted", "honeycombed", "woven". By nature, textures are repeating patterns, and as such, at least some of the featured classes should be very resilient to random cropping. For example, if the image features a "chequered" texture with a 10x10 grid of squares, then cropping out up to 89\% of the original image would still not obscure the pattern.

It is also worth noting that the mean performance decline in our experiments occurs at a more drastic rate than in \namecite{NEURIPS2022_f73c0453}. This could have been caused by differences in DA implementation, but could also stem from differences in the training data used. For example, while a human observer may conclude that the same amount of information is being removed by cropping images using the same $\alpha$, this is not the case from a model's perspective, as the absolute count of pixels removed by this operation depends on the original size of the images. This, in turn, is another possible avenue for future inquiry: experiments could be run on entirely new datasets with varying image sizes, or alternatively, on the same datasets used in this work, but with the training images upscaled.

% Possible added comment: balestriero observes "increase-plateau-decline-plateau" as alpha increases, but we actually observe "increase-plateau-decline-increase" in test accuracy in some of our classes. Worth mentioning or not?

% This is a big chunk of text but it's also the main summary and limitations/future work/discussion of our paper regarding data-centric stuff.\

% I think if we have space (it looks likely) we can add passages like this (from the original paper) referencing more exact numbers: "For example, on a resnet50 Imagenet setting, the accuracy on the “academic gown” class goes from 62% to 40% steadily as τ decreases. We defer the same experiment but using weight decay in the next section 2.4."

\begin{figure}[ht]
\centering
\includegraphics[width=1\textwidth]{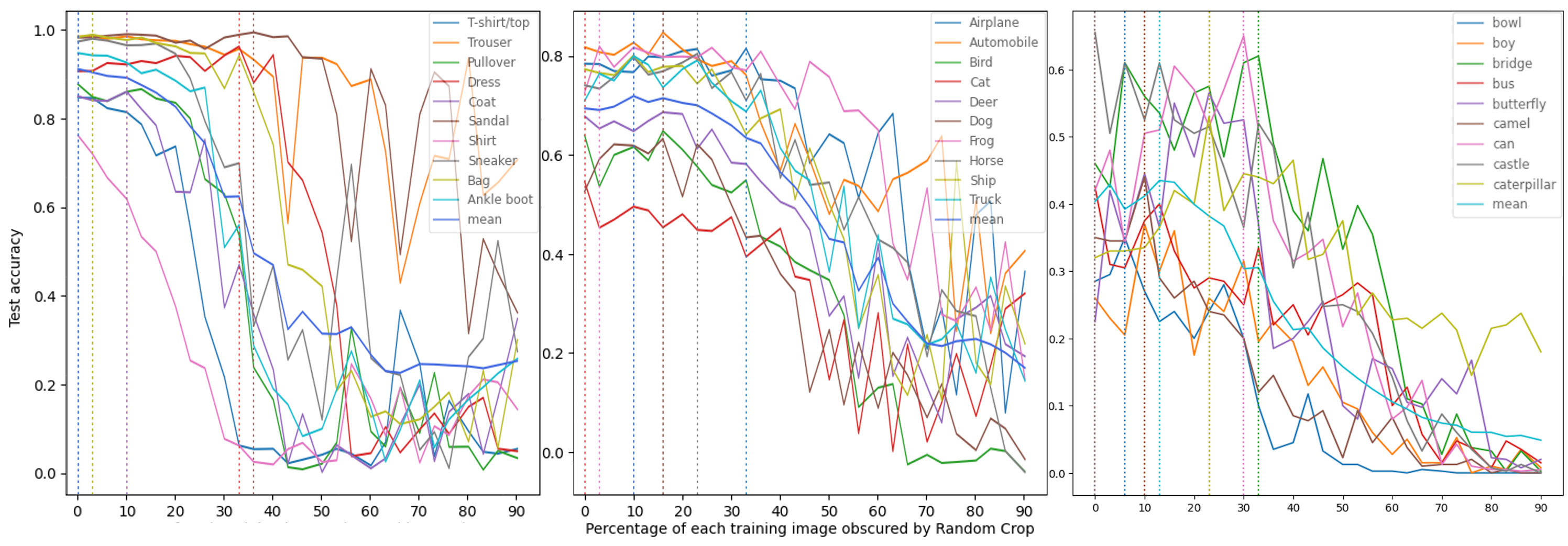}
\label{fig:resnet-flip-3data}
\caption{The results in this figure employ official ResNet50 models from Tensorflow trained from scratch on the Fashion-MNIST, CIFAR-10 \& CIFAR-100 datasets respectively, with the Random Crop and Random Horizontal Flip DA applied. All results in this figure are averaged over 4 runs. During training, the proportion of the original image obscured by the augmentation varies from 100\% to 10\%. We observe that, in line with expectations, training the same architecture on with varying random crop percentage can provide greater average test accuracy (blue) up to a certain point, but does eventually drop due to the acute negative impact some per-class accuracies experience past given crop $\alpha$ values. The vertical dotted lines denote the best test accuracy for every class. Only a subset of classes is shown for CIFAR-100 for legibility purposes.} % we can add a passage similar to Balestriero here: ""
\end{figure} 

% I think for all the results, most or all SPECIFIC numbers should go in the trash (AKA the appendix) - the tables from the original paper simply do not have room to be here. % Now that we have pretty much the complete paper written out and 1-2 pages to spare, I think this comment may or may not apply
%%%%%%%%%%%%%%%%%%%%%%%%%%%%%%%%%%%%%%%%%%%%%%
\subsection{Adding Random Horizontal Flipping Contributes To Augmentation-Induced Bias} % Can we not shorten this A.I.B. thing or omit it somehow? Thinking about it.

As part of this work's goal was to confirm the effects of DA on the bias-variance trade-off of image classification problems as seen in \namecite{NEURIPS2022_f73c0453}, we also chose to delve deeper into the specifics of the DA policy implemented by the original paper. In particular, we felt that it overlooked the possible effects its universal application of Random Horizontal Flipping (henceforth "RHF") as a supplemental DA may have introduced. In an effort to investigate this, we once again conducted a series of experiments similar to Section 2.2, this time excluding RHF. 

\begin{figure}[ht]
\centering
\includegraphics[width=1\textwidth]{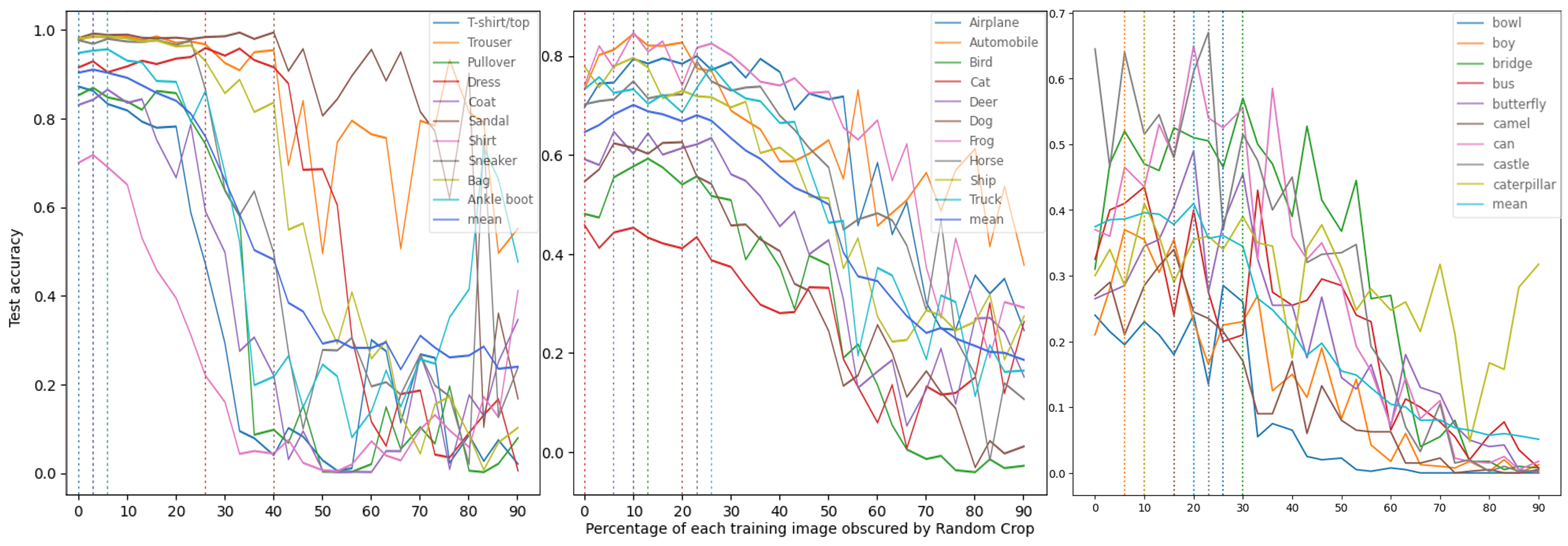}
\label{fig:resnet-noflip-3data}
\caption{The results in this figure employ official ResNet50 models from Tensorflow trained from scratch on the Fashion-MNIST, CIFAR-10 \& CIFAR-100 datasets respectively, with the Random Crop but no Random Horizontal Flip DA applied. All results in this figure are averaged over 4 runs. During training, the proportion of the original image obscured by the augmentation varies from 100\% to 10\%. We observe that while the trends from Figure 2 are generally maintained, the removal of Random Flip seems to decrease the speed at which class-specific bias manifests as $\alpha$ is increased. Only a subset of classes is shown for CIFAR-100 for legibility purposes.}
\end{figure} 

Our trials showed (see Figure 3 and appendices F, J and K) similar trends and results when compared to Section 2.1. However, as could be expected from removing a minor source of regularization such as RHF, overall mean performance was %past or present tense in results section?
marginally worse across all three datasets. In addition, it appears that the thresholds of $\alpha$ (past which overall and class-specific performances begin to fall, as well as at which best per-class and mean accuracies are reached) have generally increased - for example, "Sandal"'s best performance is up to 40\% from 36\% compared to the previous section. In this way, we see that RHF compounds with the scaling Random Cropping DA, acting as a "constant" source of additional regularization, while preserving, if accelerating, the dynamics of test set accuracies as $\alpha$ grows. With this in mind, \namecite{NEURIPS2022_f73c0453} is validated, as the conclusions reached in the work would likely not have been impacted had RHF been omitted. While not gravely consequential, this finding should serve as a reminder that caution should be exercised when chaining a plurality of data augmentations together. While such an approach is standard practice in contemporary computer vision tasks, it can rapidly increase complexity, and controlling for the influence of a given augmentation on class-specific bias may become difficult.

% This section is quite short. Since we have room - should we include some calls for future research? Unfortunately, there were no major comments on this in the thesis, so we would need to think of something extra. Should we include more exact numbers?
%%%%%%%%%%%%%%%%%%%%%%%%%%%%%%%%%%%%%%%%%%%%%%
\subsection{Alternative Architectures Have Variable Effect On Augmentation-Induced Bias} % Can we not shorten this A.I.B. thing or omit it somehow? Thinking about it.

Having carried out a data-centric analysis of DA-induced class-specific bias, we dedicated the last series of experiments (see Figure 4) to a more model-centric approach to the phenomenon. \namecite{NEURIPS2022_f73c0453} illustrates that different architectures tend to agree on the label-preserving regimes for DA($\alpha$) - in other words, that swapping ResNet50 for a different model in the above experiments would not yield a notable difference in resulting class-specific and overall performance dynamics. To validate and expand on this claim, we recreated  the experiment from Section 2.2 on the Fashion-MNIST dataset using another residual CNN, EfficientNetV2S, as well as a Vision Transformer in SWIN-Transformer.

\begin{figure}[ht]
\centering
\includegraphics[width=1\textwidth]{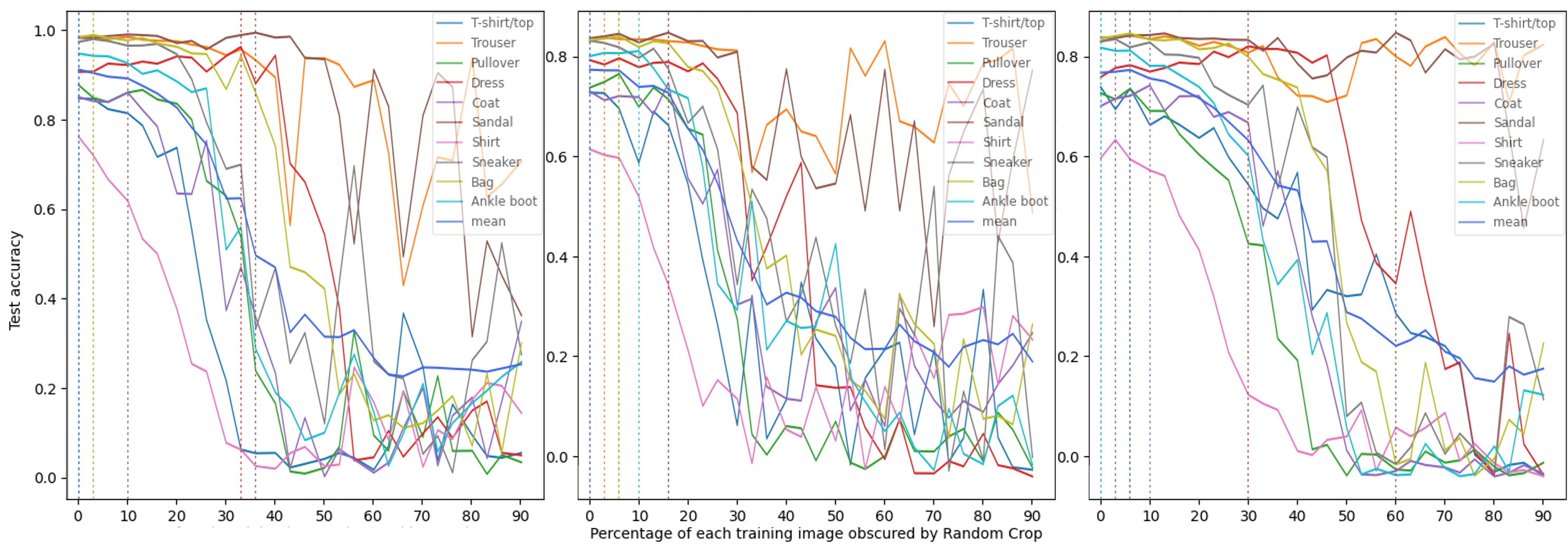}
\label{fig:fmnist-flip-3models}
\caption{The results in this figure employ official ResNet50, EfficientNetV2S and SWIN-Transformer models from Tensorflow trained from scratch on the Fashion-MNIST dataset, with the random crop and random horizontal flip DA applied. All results in this figure are averaged over 4 runs. During training, the proportion of the original image obscured by the augmentation varies from 100\% to 10\%. We observe that ResNet50 and EfficientNetV2S display very similar trends, whereas SWIN Transformer displays a noticeably delayed decline in onset of class-specific bias for certain classes.}
\end{figure} 

The "small" EfficientNetV2S \cite{DBLP:journals/corr/abs-2104-00298} architecture was selected due to being a modern and time-efficient implementation of the EfficientNet family of models. The model was downloaded, tuned and optimized using a procedure matching previous sections. %The best base performance was achieved for the EfficientNetV2S architecture trained on Fashion-MNIST with a learning rate of 0.004 and a batch size of 128, with epochs ranging from 20 to a maximum of 200.
%add specific examples: it can be seen in Figure 2 that... <fall-off is faster, as can be seen in TBA and TBA classes, but overall order and dynamics remain the same>
The EfficientNetV2S runs of this section's experiments seemed to further validate the original assumption that the phenomenon appears to be model-agnostic (though data-specific), at least for the residual family of CNNs, as the different complexity and size of the architecture did have an effect on the speed of class-specific bias occurring as the random crop $\alpha$ was increased, but the general dynamics of how performance evolved were preserved. For example, as can be seen in Figures 4, deterioration in per-class accuracy begins to occur rapidly for "Dress" at an $\alpha$ value of 30\% for EfficientNetV2S, in contrast to 40\% for ResNet50. For a full summary of the diverse $\alpha$ at which each class and mean test set performance reach their peak, see Appendix I.

With the previous section confirming the model-agnostic nature of the phenomenon as it pertains to residual CNNs, the final trial followed up by evaluating the performance of a SWIN Transformer, as the architecture is composed in a fundamentally different way, and can be adjusted to perceive images with a finer level of granularity via the patch size parameter. 
The model was prepared similarly to previous sections. %The best base performance was achieved for the SWIN Transformer architecture trained on Fashion-MNIST with a learning rate of 0.0008 and a batch size of 128, with epochs ranging from 20 to a maximum of 200 and a patch size of 2x2. %give some concrete examples from the plot - use thesis for detailed description of trends
As visible in Figure 4 and Appendix L, while the use of a Vision Transformer model did not entirely avert the overall class-specific bias trend as random cropping was applied more aggressively, it severely delayed its heavier impacts, and even noticeably changed the behavior of certain classes (for example, maintaining the stability of the top performers in "Trouser" and "Sandal" even for very high values of $\alpha$, while delaying "Dress" and "Coat"'s rapid deterioration in accuracy to 47\% and 30\% from 40\% and 23\%, respectively) in effect making the model-dataset combination more robust against the potential negative trade-offs data augmentation can bring. Despite this, and being similar in terms of computational requirements, its best performance achieved in practice was marginally lower than that of ResNet50 
%(0.901 vs 0.912) % include?
. While it does not outright disagree with \namecite{NEURIPS2022_f73c0453}'s model-agnostic proposition, the implications of this result are such that architectures for computer vision tasks can and should be chosen not just by best overall test set performance, but also based on other merits, such as robustness to bias and label loss resulting from aggressive data augmentations, especially in cases where such augmentations are expected to be applied en masse or without intensive monitoring, such as in MLOps systems with regular retraining.

While the above experiment does serve to illustrate the potential merit of applying alternative architectures in the context of regulating DA-induced class-specific bias, broader research is likely warranted. A natural direction of such an extension may be to continue with applying different families and instances of image-classifying neural network architectures to the task. One option is to expand the research utilizing SWIN Transformer by testing it on a further plurality of datasets with a wider variety of patch sizes, as the concepts of regulating the granularity with which the model learns to see details in images and the random cropping DA are conceptually adjacent. Following this, exploring the same problem with different Vision Transformers, such as the more lightweight MobileViT \cite{DBLP:journals/corr/abs-2110-02178} or Google's family of large ViT models \cite{Zhai_2022_CVPR} is another step in that direction. Another possible direction to explore the model-specific degree of the phenomenon is to investigate how it pertains to Capsule Networks, first described in \namecite{sabour2017dynamic}, as the architecture, once again, vastly differs from both Residual CNNs and Vision Transformers, and is generally considered to be less susceptible to variance from image transformations. This family of models is also very demanding to train, and would require a greater investment of resources or smaller scope than what our work had. As CapsNets are, at the time of writing, a very active area of research \cite{KWABENAPATRICK20221295}, we recommend this as a very valuable line of inquiry. 

\section{Conclusion and Limitations}

This study extends the analysis initiated by \namecite{NEURIPS2022_f73c0453}, focusing on the impact of data augmentations, particularly Random Crop, on class-specific bias in image classification models. Our contributions are multi-faceted, addressing the need for a more nuanced understanding of DA's effects across different contexts.

We empirically demonstrate that DA-induced class-specific biases are not exclusive to ImageNet but also affect datasets with distinct characteristics, such as Fashion-MNIST and CIFAR. These datasets, featuring significantly fewer and smaller-sized images, some of which are monochrome, provide a broader canvas to assess DA's impact. This variation in dataset characteristics allowed us to explore how DA-induced biases manifest in environments markedly different from ImageNet, thus broadening the scope of understanding regarding DA's implications.

By incorporating additional deep neural network architectures like EfficientNetV2S (a residual model) and SWIN Vision Transformer (a non-residual, patch-based model), we delve into the model-agnostic proposition of class-specific DA-induced bias. Our findings reveal that while the phenomenon extends to residual models, alternative architectures such as Vision Transformers exhibit a varying degree of robustness or altered dynamics in response to DA, suggesting a potential strategy for mitigating class-specific biases through architectural selection.

We offer a detailed methodology for "data augmentation robustness scouting," refining the initial concept proposed by \namecite{NEURIPS2022_f73c0453}. This step-by-step approach aims at a more efficient, resource-sensitive examination of DA's effects, facilitating the identification and mitigation of class-specific biases in the model design phase. By applying this methodology, we not only validate previous findings but also present a practical framework for future studies and model development efforts.

Our study, while highlighting the aforementioned contributions, acknowledges its scope limitations concerning the variety of architectures and datasets examined. Future work is encouraged to explore a broader array of computer vision models and data characteristics, potentially unveiling novel insights into DA's nuanced effects on model performance and bias. This endeavor aims not only to deepen our understanding of DA's impact across different settings but also to contribute to the development of more equitable and effective computer vision systems.

%%%%%%%%%%%%%%%%%%%%%%%%%%%%%%%%%%%%%%%%%%%%%%
%%%%%%%%%%%%%%%%%%%%%%%%%%%%%%%%%%%%%%%%%%%%%%
\pagebreak
\newpage
\bibliography{compling_style}

\newpage
\appendix

\section*{Appendices}

\appendixsection{Image dimensions (in pixels) off training images after being randomly cropped and before being resized} \label{apx:first}

[32x32, 31x31, 30x30,

29x29, 28x28, 27x27,

26x26, 25x25, 24x24,

22x22, 21x21, 20x20,
 
19x19, 18x18, 17x17,

16x16, 15x15, 14x14,
 
13x13, 12x12, 11x11,

10x10, 9x9, 8x8, 

6x6,5x5, 4x4, 3x3]

\appendixsection{Dataset samples corresponding to the Fashion-MNIST segment used in training} \label{apx:second}

\begin{figure}[H]
\centering
\includegraphics[width=1\textwidth]{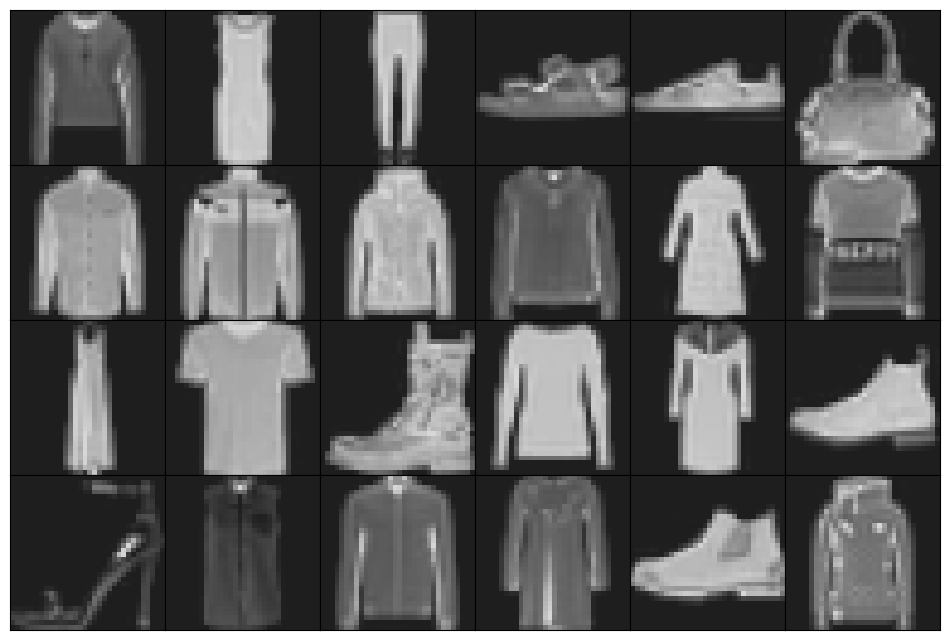}
\end{figure} 
\pagebreak
\appendixsection{Dataset samples corresponding to the CIFAR-10 segment used in training} 
\label{apx:third}

\begin{figure}[H]
\centering
\includegraphics[width=1\textwidth]{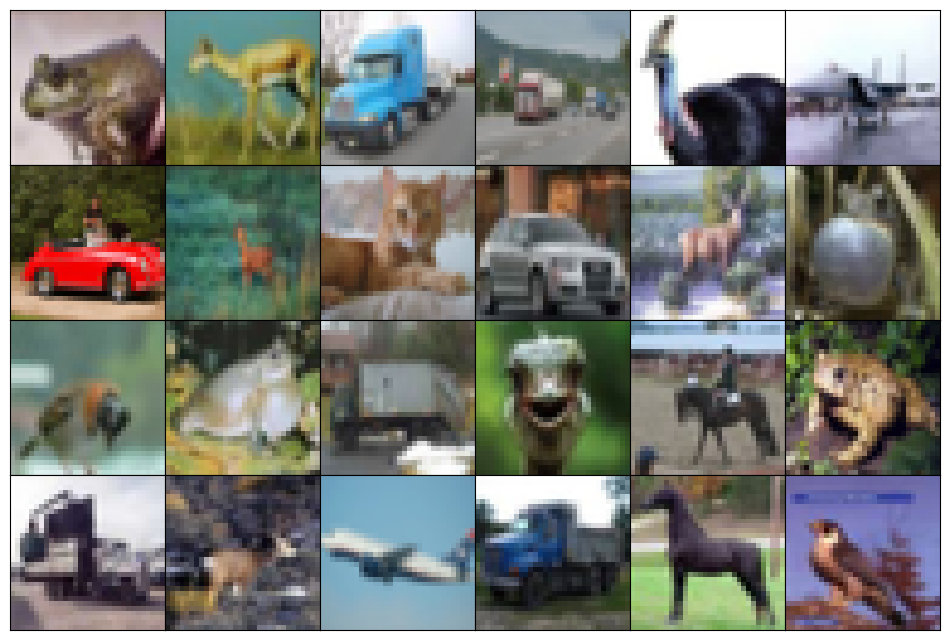}
\end{figure} 
\pagebreak
\appendixsection{Dataset samples corresponding to the CIFAR-100 segment used in training} 
\label{apx:fourth}

\begin{figure}[H]
\centering
\includegraphics[width=1\textwidth]{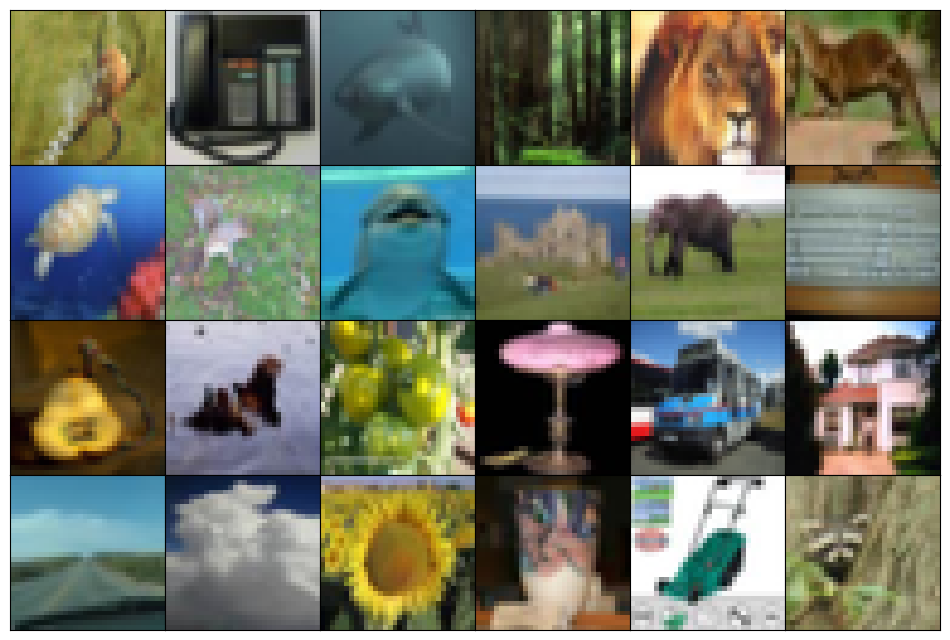}
\end{figure} 
\pagebreak

\appendixsection{Full collection of class accuracy plots for CIFAR-100} 
\label{apx:fifth}

\begin{figure}[H]
     \centering
     \begin{subfigure}[b]{0.3\textwidth}
         \centering
         \includegraphics[width=\textwidth]{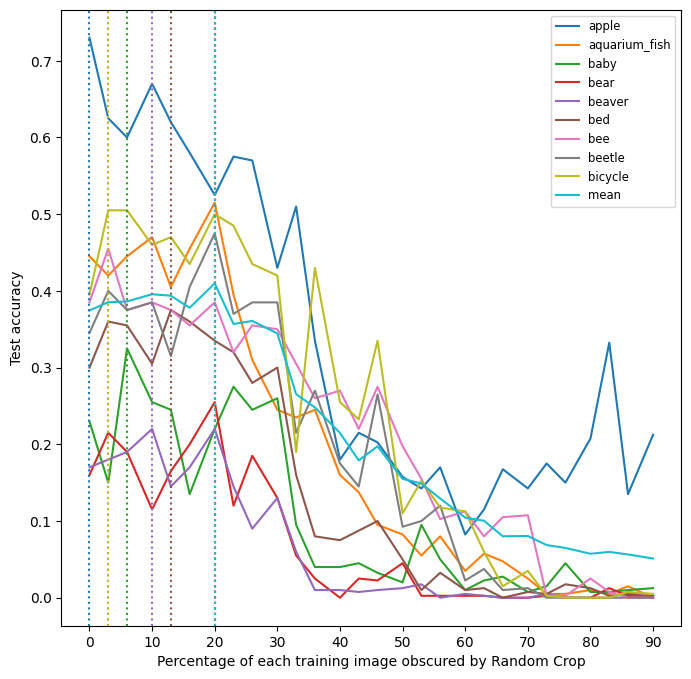}
     \end{subfigure}
     \hfill
     \begin{subfigure}[b]{0.3\textwidth}
         \centering
         \includegraphics[width=\textwidth]{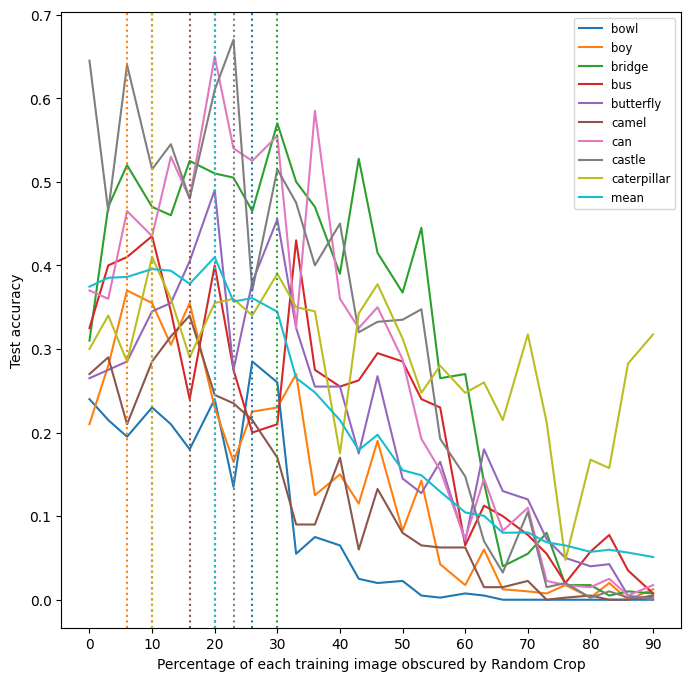}
     \end{subfigure}
     \hfill
     \begin{subfigure}[b]{0.3\textwidth}
         \centering
         \includegraphics[width=\textwidth]{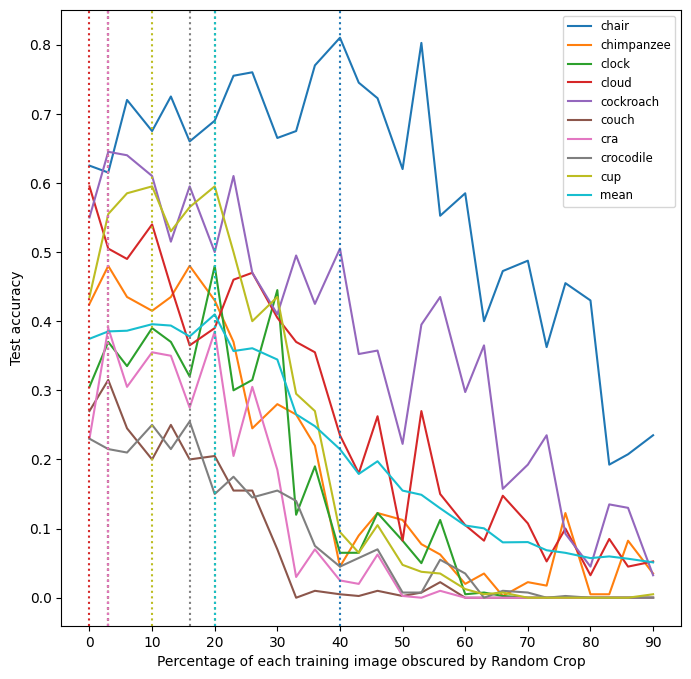}
     \end{subfigure}
\end{figure}

\begin{figure}[H]
     \centering
     \begin{subfigure}[b]{0.3\textwidth}
         \centering
         \includegraphics[width=\textwidth]{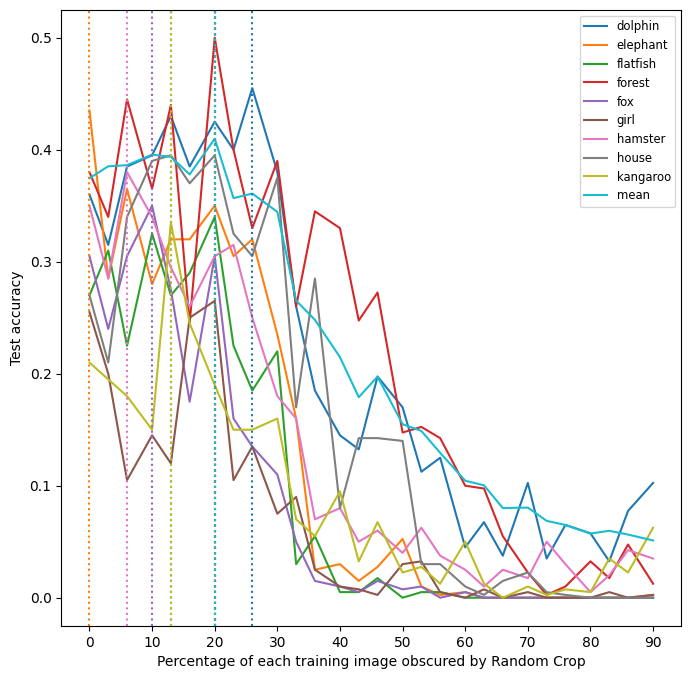}
     \end{subfigure}
     \hfill
     \begin{subfigure}[b]{0.3\textwidth}
         \centering
         \includegraphics[width=\textwidth]{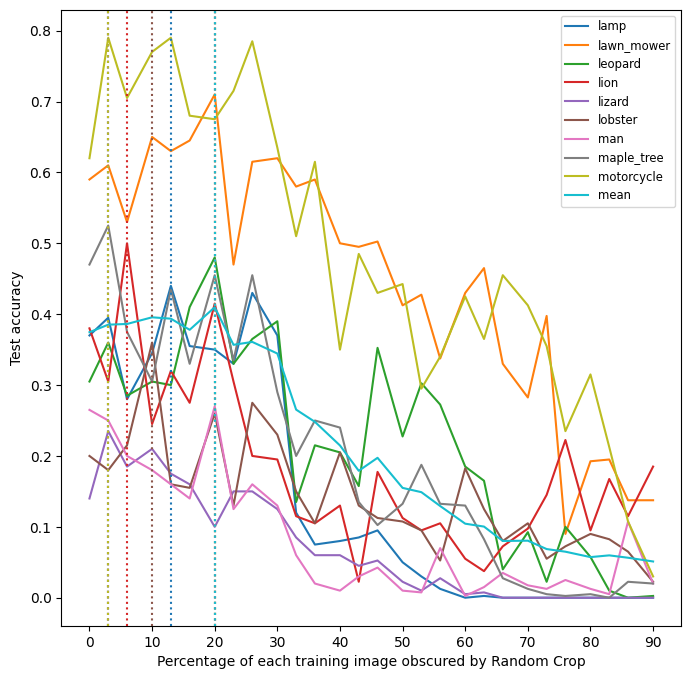}
     \end{subfigure}
     \hfill
     \begin{subfigure}[b]{0.3\textwidth}
         \centering
         \includegraphics[width=\textwidth]{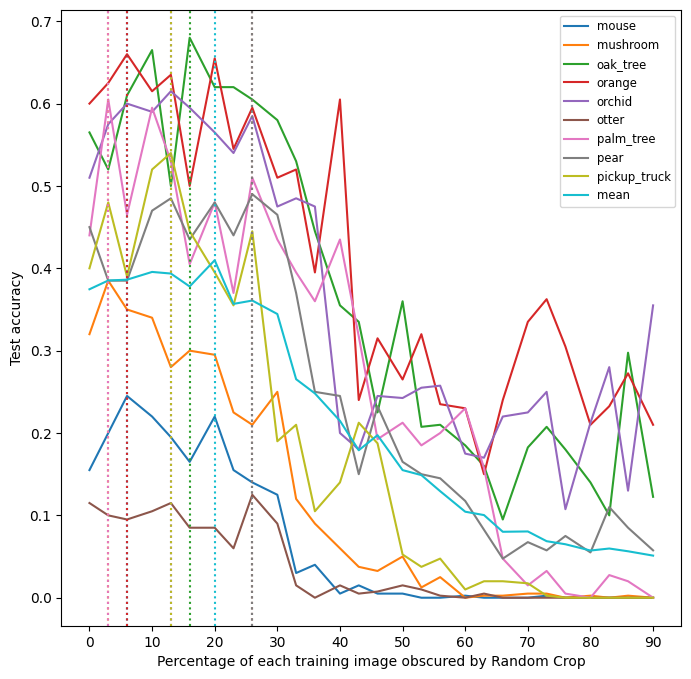}
     \end{subfigure}
\end{figure}

\begin{figure}[H]
     \centering
     \begin{subfigure}[b]{0.3\textwidth}
         \centering
         \includegraphics[width=\textwidth]{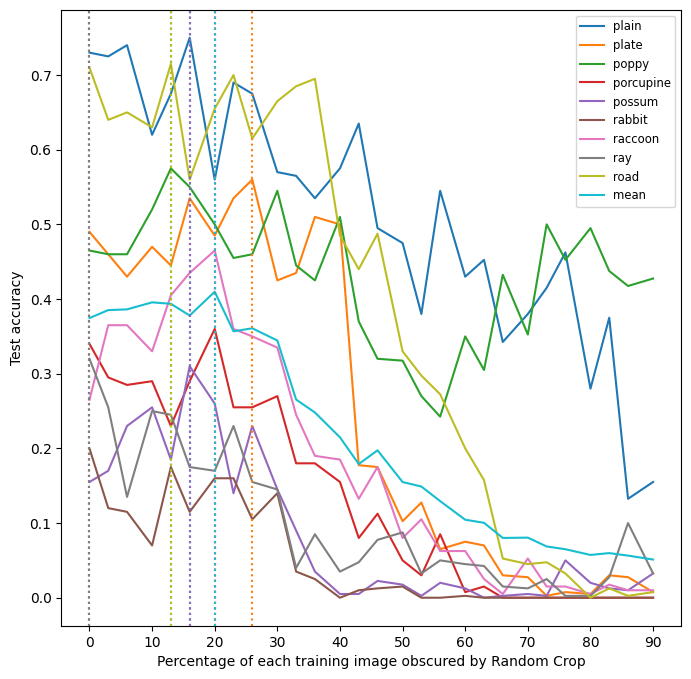}
     \end{subfigure}
     \hfill
     \begin{subfigure}[b]{0.3\textwidth}
         \centering
         \includegraphics[width=\textwidth]{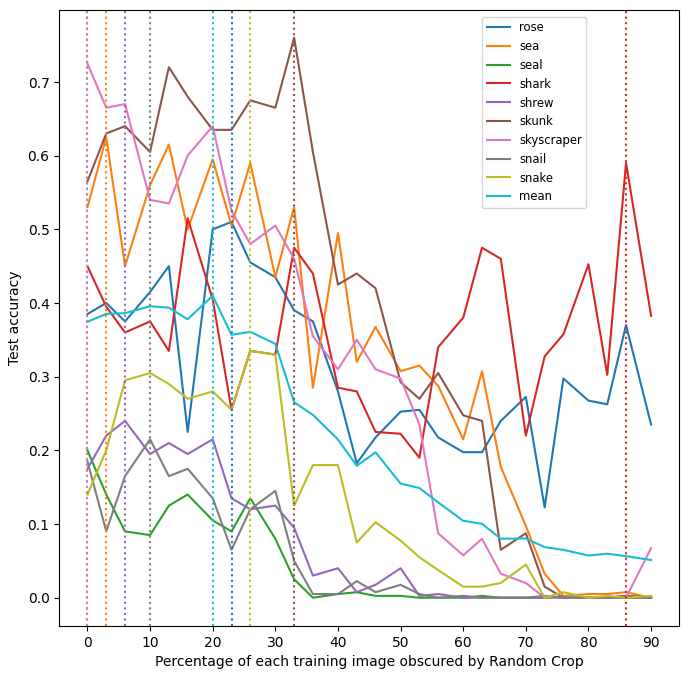}
     \end{subfigure}
     \hfill
     \begin{subfigure}[b]{0.3\textwidth}
         \centering
         \includegraphics[width=\textwidth]{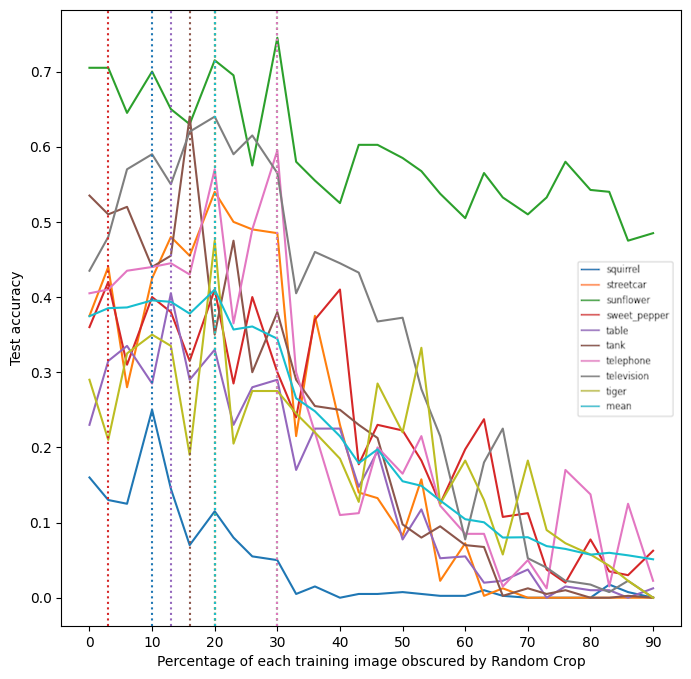}
     \end{subfigure}
     \hfill
\end{figure}

\begin{figure}[H]
    \centering
    \begin{subfigure}[b]{0.3\textwidth}
         \centering
         \includegraphics[width=\textwidth]{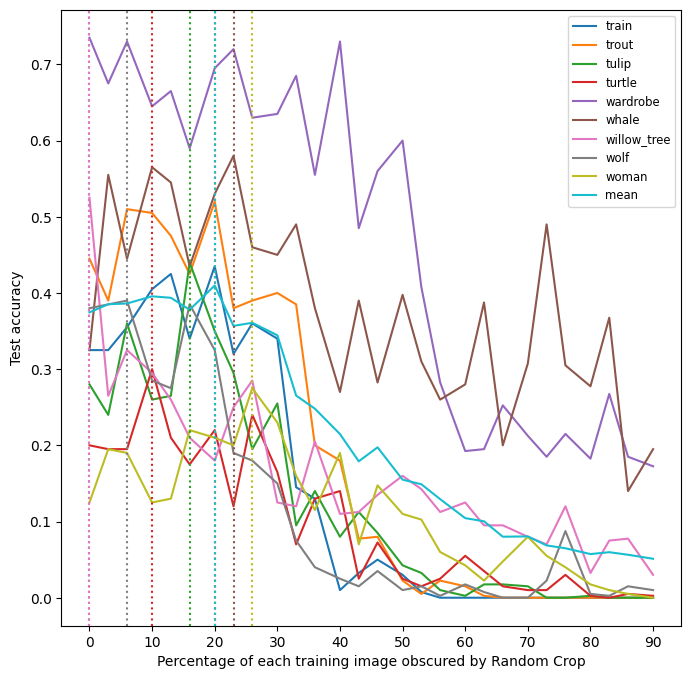}
     \end{subfigure}
     \begin{subfigure}[b]{0.65\textwidth}
         \centering
         \includegraphics[width=\textwidth]{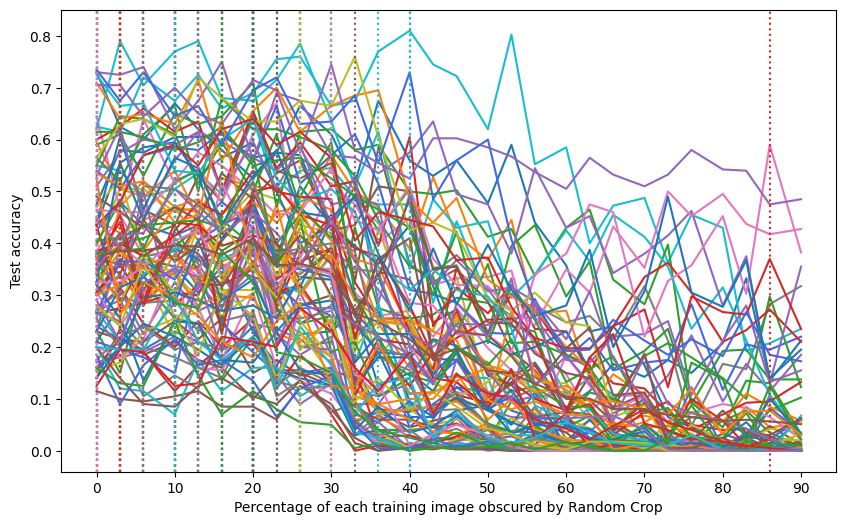}
     \end{subfigure}
\subcaption{The results in all figures employ official ResNet50 models from Tensorflow trained from scratch on the CIFAR-100 dataset with random crop data augmentation applied. All results in this figure are averaged over 4 runs. During training, the proportion of the original image obscured by the augmentation varies from 100\% to 10\%. We observe  The vertical dotted lines denote the best test accuracy for every class.}
\end{figure}

\begin{figure}[H]
     \centering
     \begin{subfigure}[b]{0.3\textwidth}
         \centering
         \includegraphics[width=\textwidth]{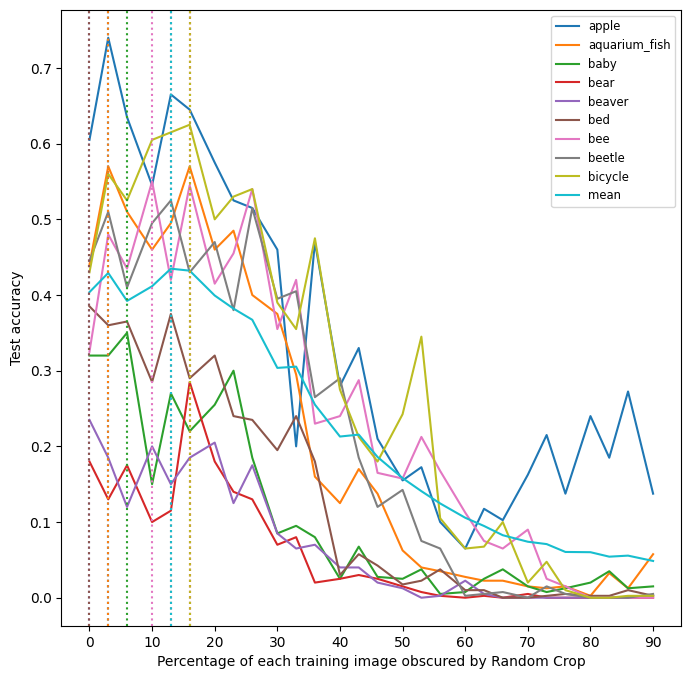}
     \end{subfigure}
     \hfill
     \begin{subfigure}[b]{0.3\textwidth}
         \centering
         \includegraphics[width=\textwidth]{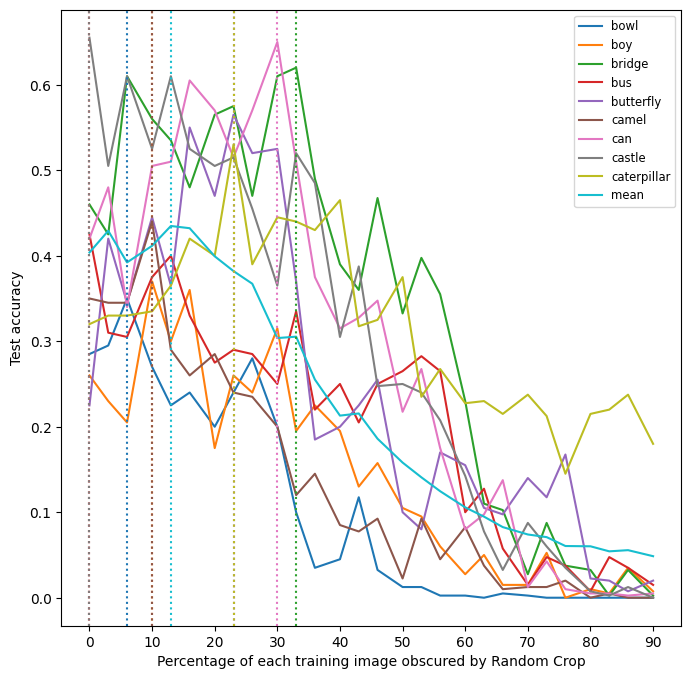}
     \end{subfigure}
     \hfill
     \begin{subfigure}[b]{0.3\textwidth}
         \centering
         \includegraphics[width=\textwidth]{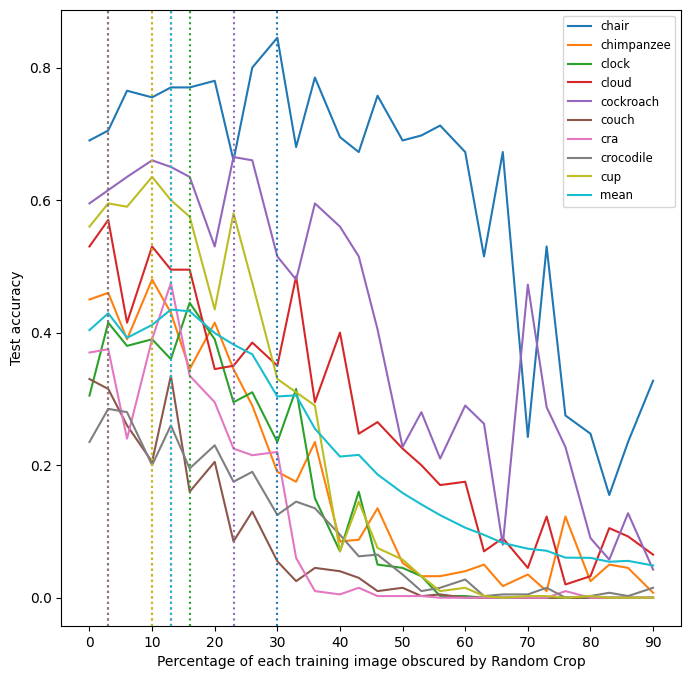}
     \end{subfigure}
\end{figure}

\begin{figure}[H]
     \centering
     \begin{subfigure}[b]{0.3\textwidth}
         \centering
         \includegraphics[width=\textwidth]{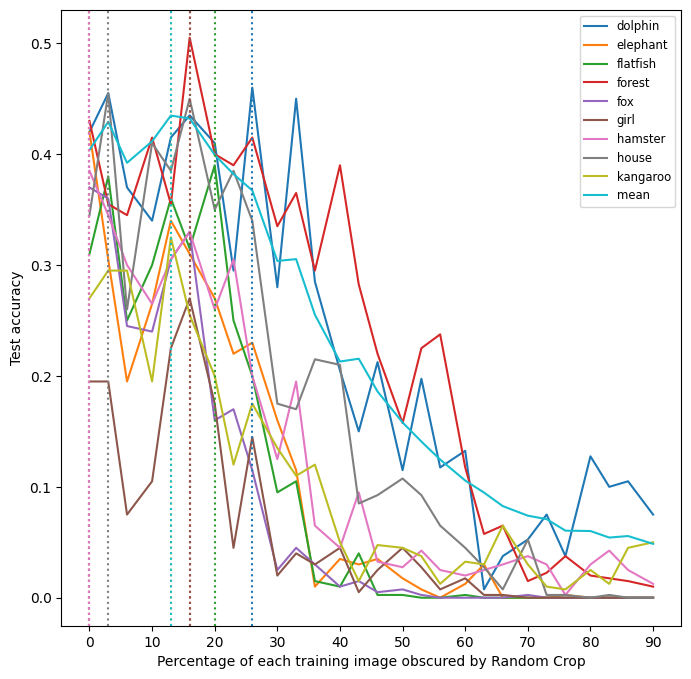}
     \end{subfigure}
     \hfill
     \begin{subfigure}[b]{0.3\textwidth}
         \centering
         \includegraphics[width=\textwidth]{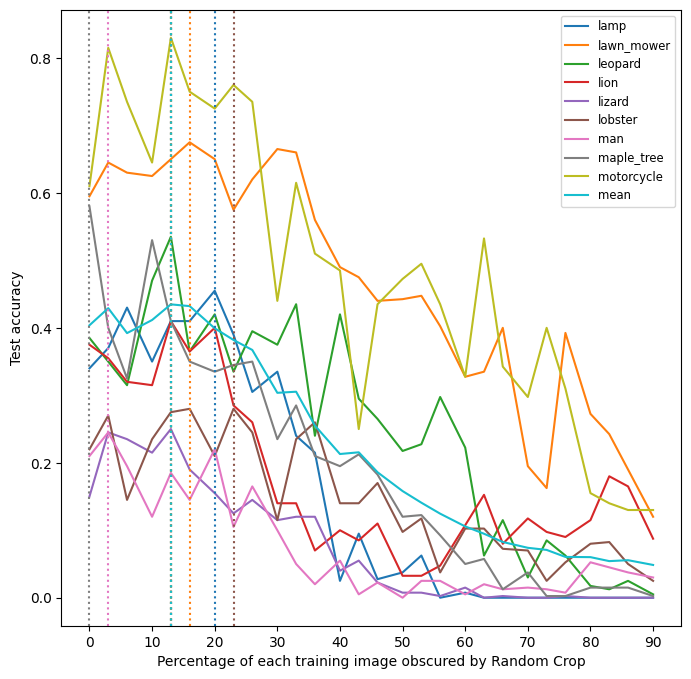}
     \end{subfigure}
     \hfill
     \begin{subfigure}[b]{0.3\textwidth}
         \centering
         \includegraphics[width=\textwidth]{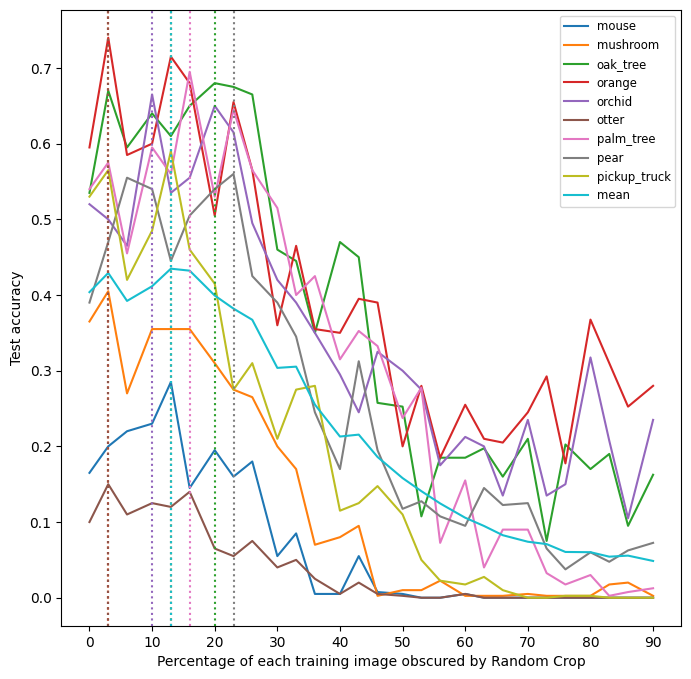}
     \end{subfigure}
\end{figure}

\begin{figure}[H]
     \centering
     \begin{subfigure}[b]{0.3\textwidth}
         \centering
         \includegraphics[width=\textwidth]{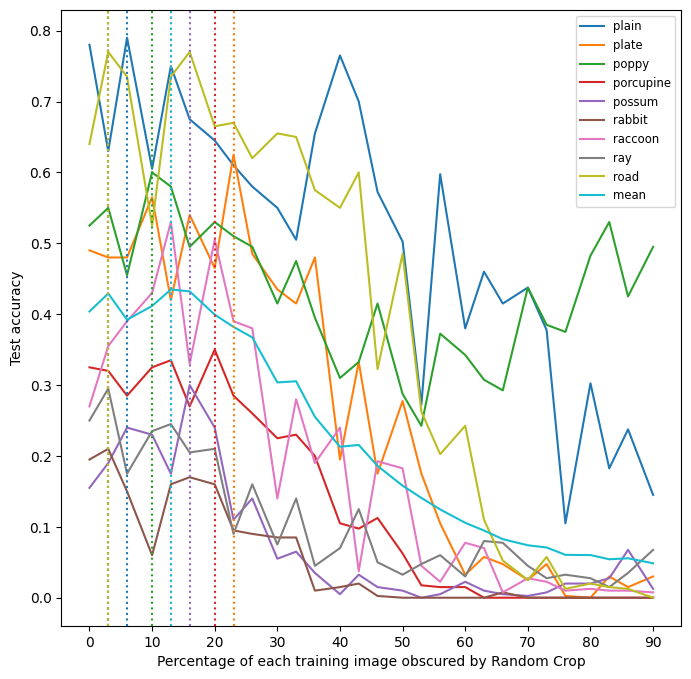}
     \end{subfigure}
     \hfill
     \begin{subfigure}[b]{0.3\textwidth}
         \centering
         \includegraphics[width=\textwidth]{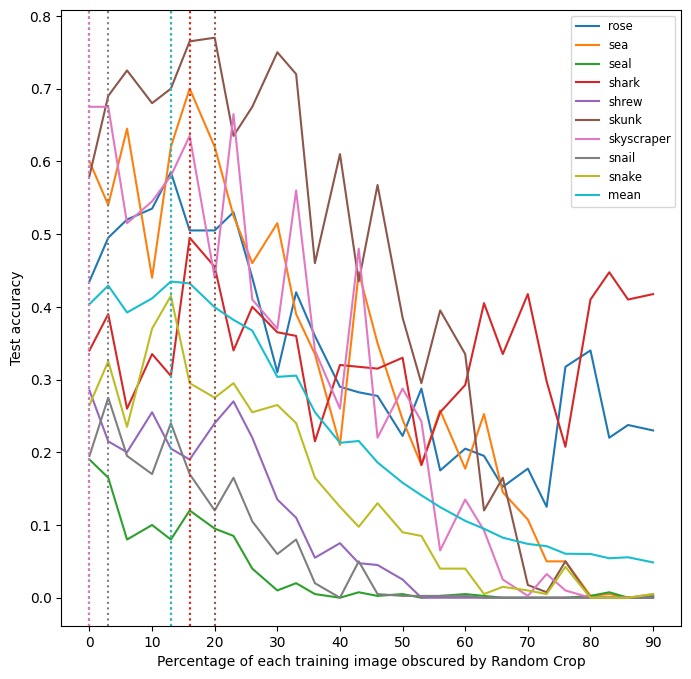}
     \end{subfigure}
     \hfill
     \begin{subfigure}[b]{0.3\textwidth}
         \centering
         \includegraphics[width=\textwidth]{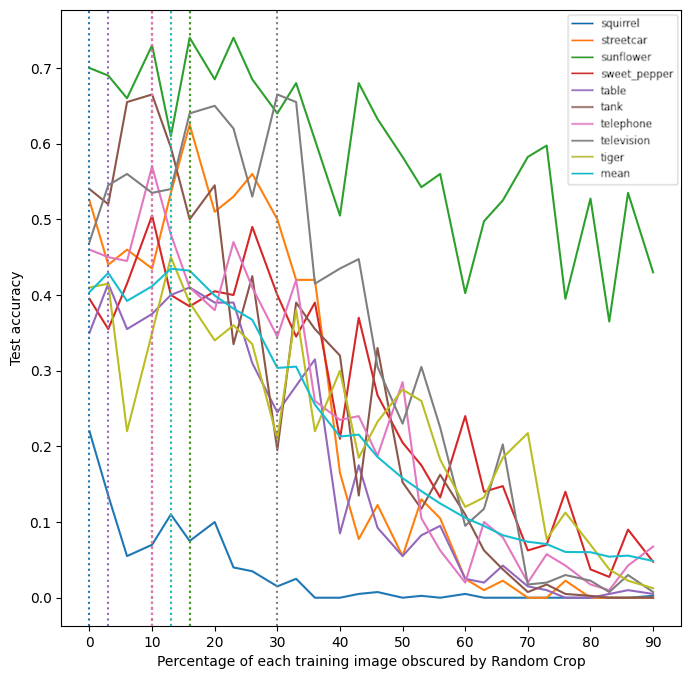}
     \end{subfigure}
     \hfill
\end{figure}

\begin{figure}[H]
    \centering
    \begin{subfigure}[b]{0.3\textwidth}
         \centering
         \includegraphics[width=\textwidth]{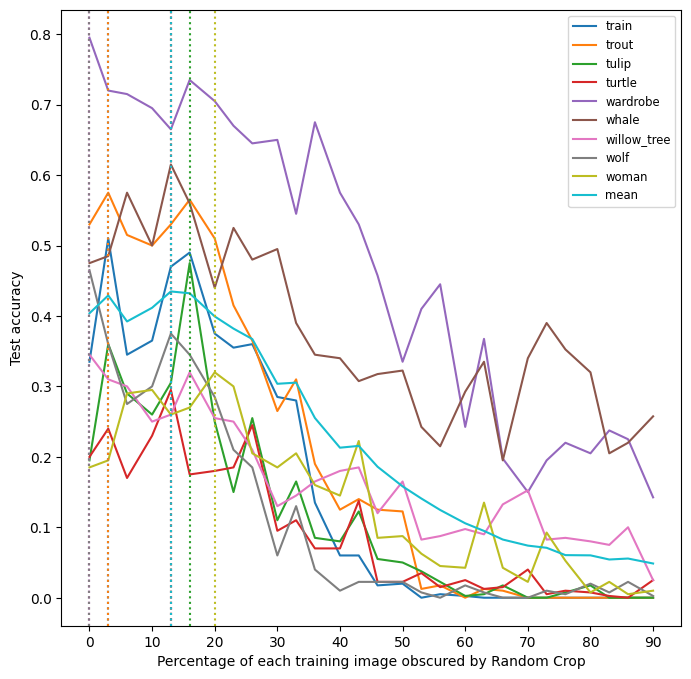}
     \end{subfigure}
     \begin{subfigure}[b]{0.65\textwidth}
         \centering
         \includegraphics[width=\textwidth]{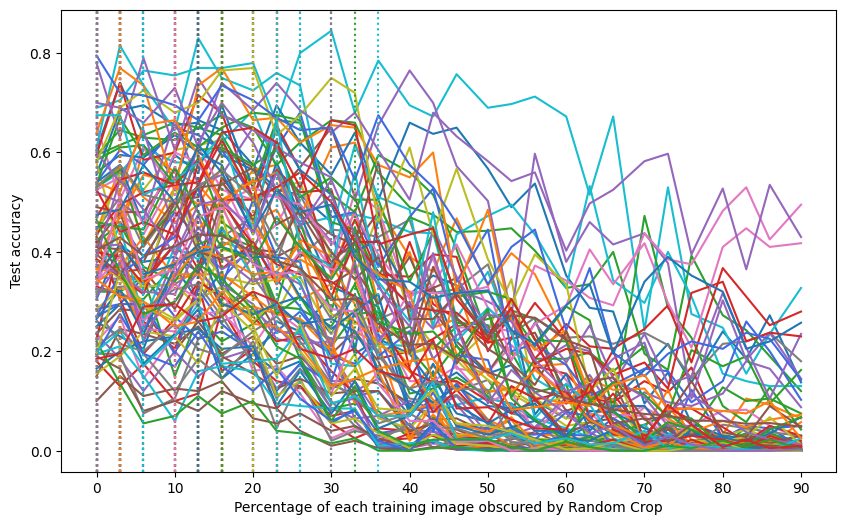}
     \end{subfigure}
\subcaption{The results in all figures employ official ResNet50 models from Tensorflow trained from scratch on the CIFAR-100 dataset with random crop and random horizontal flip data augmentations applied. All results in this figure are averaged over 4 runs. During training, the proportion of the original image obscured by the augmentation varies from 100\% to 10\%. We observe  The vertical dotted lines denote the best test accuracy for every class.}
\end{figure}
\pagebreak

\newpage
\appendixsection{Full collection of best test performances for CIFAR-100} 
\label{apx:sixth}

\textbf{Without Random Horizontal Flip:}

\begin{longtable}{lllllll}
              & chair     & motorcycle     & skunk       & plain         & sunflower    & wardrobe    \\
Random Crop $\alpha$ & 40        & 3              & 33          & 16            & 30           & 0           \\
Test Accuracy & 0.81      & 0.79           & 0.76        & 0.75          & 0.745        & 0.735       \\
              & apple     & skyscraper     & road        & lawn\_mower   & oak\_tree    & keyboard    \\
Random Crop $\alpha$ & 0         & 0              & 13          & 20            & 16           & 36          \\
Test Accuracy & 0.73      & 0.725          & 0.715       & 0.71          & 0.68         & 0.675       \\
              & castle    & mountain       & orange      & can           & cockroach    & television  \\
Random Crop $\alpha$ & 23        & 23             & 6           & 20            & 3            & 20          \\
Test Accuracy & 0.67      & 0.665          & 0.66        & 0.65          & 0.645        & 0.64        \\
              & bottle    & tank           & sea         & rocket        & orchid       & palm\_tree  \\
Random Crop $\alpha$ & 20        & 16             & 3           & 16            & 13           & 3           \\
Test Accuracy & 0.64      & 0.64           & 0.625       & 0.625         & 0.615        & 0.605       \\
              & cloud     & telephone      & cup         & shark         & whale        & poppy       \\
Random Crop $\alpha$ & 0         & 30             & 10          & 86            & 23           & 13          \\
Test Accuracy & 0.595     & 0.595          & 0.595       & 0.59          & 0.58         & 0.575       \\
              & bridge    & plate          & streetcar   & pickup\_truck & willow\_tree & maple\_tree \\
Random Crop $\alpha$ & 30        & 26             & 20          & 13            & 0            & 3           \\
Test Accuracy & 0.57      & 0.56           & 0.54        & 0.54          & 0.525        & 0.525       \\
              & trout     & aquarium\_fish & rose        & worm          & bicycle      & lion        \\
Random Crop $\alpha$ & 20        & 20             & 23          & 10            & 3            & 6           \\
Test Accuracy & 0.52      & 0.515          & 0.51        & 0.505         & 0.505        & 0.5         \\
              & forest    & butterfly      & pear        & tractor       & chimpanzee   & clock       \\
Random Crop $\alpha$ & 20        & 20             & 26          & 40            & 3            & 20          \\
Test Accuracy & 0.5       & 0.49           & 0.49        & 0.485         & 0.48         & 0.48        \\
              & leopard   & beetle         & tiger       & raccoon       & spider       & pine\_tree  \\
Random Crop $\alpha$ & 20        & 20             & 20          & 20            & 26           & 3           \\
Test Accuracy & 0.48      & 0.475          & 0.475       & 0.465         & 0.465        & 0.455       \\
              & dolphin   & bee            & tulip       & lamp          & bus          & train       \\
Random Crop $\alpha$ & 26        & 3              & 16          & 13            & 10           & 20          \\
Test Accuracy & 0.455     & 0.455          & 0.44        & 0.44          & 0.435        & 0.435       \\
              & elephant  & sweet\_pepper  & caterpillar & mean          & table        & cattle      \\
Random Crop $\alpha$ & 0         & 3              & 10          & 20            & 13           & 30          \\
Test Accuracy & 0.435     & 0.42           & 0.41        & 0.4098        & 0.405        & 0.395       \\
              & house     & cra            & wolf        & mushroom      & dinosaur     & hamster     \\
Random Crop $\alpha$ & 13        & 3              & 6           & 3             & 10           & 6           \\
Test Accuracy & 0.395     & 0.39           & 0.39        & 0.385         & 0.38         & 0.38        \\
              & bed       & boy            & porcupine   & lobster       & fox          & camel       \\
Random Crop $\alpha$ & 13        & 6              & 20          & 10            & 10           & 16          \\
Test Accuracy & 0.375     & 0.37           & 0.36        & 0.36          & 0.35         & 0.34        \\
              & flatfish  & snake          & kangaroo    & baby          & ray          & couch       \\
Random Crop $\alpha$ & 20        & 26             & 13          & 6             & 0            & 3           \\
Test Accuracy & 0.34      & 0.335          & 0.335       & 0.325         & 0.32         & 0.315       \\
              & possum    & turtle         & bowl        & woman         & man          & girl        \\
Random Crop $\alpha$ & 16        & 10             & 26          & 26            & 20           & 20          \\
Test Accuracy & 0.31      & 0.3            & 0.285       & 0.275         & 0.27         & 0.265       \\
              & crocodile & bear           & squirrel    & mouse         & shrew        & lizard      \\
Random Crop $\alpha$ & 16        & 20             & 10          & 6             & 6            & 3           \\
Test Accuracy & 0.255     & 0.255          & 0.25        & 0.245         & 0.24         & 0.235       \\
              & beaver    & snail          & rabbit      & seal          & otter        & ideal       \\
Random Crop $\alpha$ & 10        & 10             & 0           & 0             & 26           & N/A          \\
Test Accuracy & 0.22      & 0.215          & 0.2         & 0.2           & 0.125        & 0.47405    
\end{longtable}

\textbf{With Random Horizontal Flip}

\begin{longtable}{lllllll}
              & chair      & motorcycle    & wardrobe      & plain          & skunk      & road        \\
Random Crop $\alpha$ & 30         & 13            & 0             & 6              & 20         & 3           \\
Test Accuracy & 0.845      & 0.83          & 0.795         & 0.79           & 0.77       & 0.77        \\
              & apple      & sunflower     & orange        & sea            & palm\_tree & keyboard    \\
Random Crop $\alpha$ & 3          & 16            & 3             & 16             & 16         & 6           \\
Test Accuracy & 0.74       & 0.74          & 0.74          & 0.7            & 0.695      & 0.695       \\
              & bottle     & mountain      & oak\_tree     & lawn\_mower    & skyscraper & orchid      \\
Random Crop $\alpha$ & 20         & 26            & 20            & 16             & 0          & 10          \\
Test Accuracy & 0.685      & 0.68          & 0.68          & 0.675          & 0.675      & 0.665       \\
              & television & cockroach     & tank          & castle         & can        & cup         \\
Random Crop $\alpha$ & 30         & 23            & 10            & 0              & 30         & 10          \\
Test Accuracy & 0.665      & 0.665         & 0.665         & 0.655          & 0.65       & 0.635       \\
              & plate      & bicycle       & streetcar     & bridge         & whale      & rocket      \\
Random Crop $\alpha$ & 23         & 16            & 16            & 33             & 13         & 6           \\
Test Accuracy & 0.625      & 0.625         & 0.625         & 0.62           & 0.615      & 0.615       \\
              & poppy      & pickup\_truck & rose          & pine\_tree     & tractor    & maple\_tree \\
Random Crop $\alpha$ & 10         & 13            & 13            & 13             & 36         & 0           \\
Test Accuracy & 0.6        & 0.59          & 0.585         & 0.58           & 0.58       & 0.58        \\
              & trout      & telephone     & cloud         & aquarium\_fish & butterfly  & pear        \\
Random Crop $\alpha$ & 3          & 10            & 3             & 3              & 23         & 23          \\
Test Accuracy & 0.575      & 0.57          & 0.57          & 0.57           & 0.565      & 0.56        \\
              & worm       & bee           & leopard       & caterpillar    & raccoon    & beetle      \\
Random Crop $\alpha$ & 23         & 10            & 13            & 23             & 13         & 13          \\
Test Accuracy & 0.56       & 0.55          & 0.535         & 0.53           & 0.53       & 0.525       \\
              & train      & forest        & sweet\_pepper & spider         & shark      & chimpanzee  \\
Random Crop $\alpha$ & 3          & 16            & 10            & 3              & 16         & 10          \\
Test Accuracy & 0.51       & 0.505         & 0.505         & 0.495          & 0.495      & 0.48        \\
              & tulip      & cra           & wolf          & dolphin        & house      & lamp        \\
Random Crop $\alpha$ & 16         & 13            & 0             & 26             & 3          & 20          \\
Test Accuracy & 0.475      & 0.475         & 0.465         & 0.46           & 0.455      & 0.455       \\
              & tiger      & clock         & camel         & mean           & bus        & elephant    \\
Random Crop $\alpha$ & 13         & 16            & 10            & 13             & 0          & 0           \\
Test Accuracy & 0.45       & 0.445         & 0.44          & 0.4348         & 0.425      & 0.42        \\
              & table      & snake         & lion          & cattle         & mushroom   & flatfish    \\
Random Crop $\alpha$ & 3          & 13            & 13            & 16             & 3          & 20          \\
Test Accuracy & 0.415      & 0.415         & 0.41          & 0.405          & 0.405      & 0.39        \\
              & bed        & hamster       & boy           & fox            & dinosaur   & bowl        \\
Random Crop $\alpha$ & 0          & 0             & 10            & 0              & 0          & 6           \\
Test Accuracy & 0.385      & 0.385         & 0.37          & 0.37           & 0.365      & 0.35        \\
              & baby       & porcupine     & willow\_tree  & couch          & kangaroo   & woman       \\
Random Crop $\alpha$ & 6          & 20            & 0             & 13             & 13         & 20          \\
Test Accuracy & 0.35       & 0.35          & 0.345         & 0.335          & 0.325      & 0.32        \\
              & possum     & ray           & turtle        & crocodile      & shrew      & bear        \\
Random Crop $\alpha$ & 16         & 3             & 13            & 3              & 0          & 16          \\
Test Accuracy & 0.3        & 0.295         & 0.295         & 0.285          & 0.285      & 0.285       \\
              & mouse      & lobster       & snail         & girl           & lizard     & man         \\
Random Crop $\alpha$ & 13         & 23            & 3             & 16             & 13         & 3           \\
Test Accuracy & 0.285      & 0.28          & 0.275         & 0.27           & 0.25       & 0.245       \\
              & beaver     & squirrel      & rabbit        & seal           & otter      & ideal       \\
Random Crop $\alpha$ & 0          & 0             & 3             & 0              & 3          & N/A       \\
Test Accuracy & 0.235      & 0.22          & 0.21          & 0.19           & 0.15       & 0.49985    
\end{longtable}

\appendixsection{Per-class and overall test set performances samples for the Fashion-MNIST + ResNet50 + Random Cropping + Random Horizontal Flip experiment} \label{apx:eight}

\begin{figure}[H]
\centering
\includegraphics[width=1\textwidth]{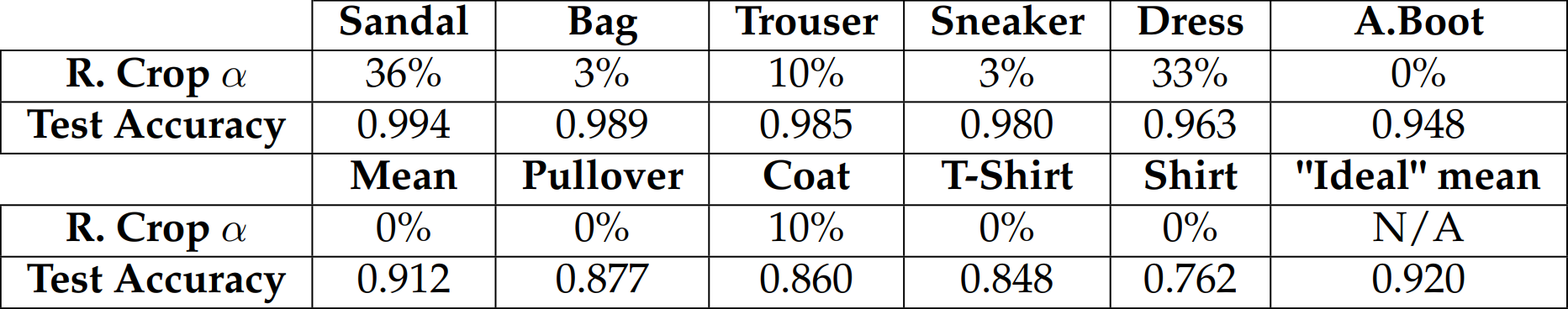}
\end{figure} 

\appendixsection{Per-class and overall test set performances samples for the CIFAR-10 + ResNet50 + Random Cropping + Random Horizontal Flip experiment} \label{apx:twelfth}

\begin{figure}[H]
\centering
\includegraphics[width=1\textwidth]{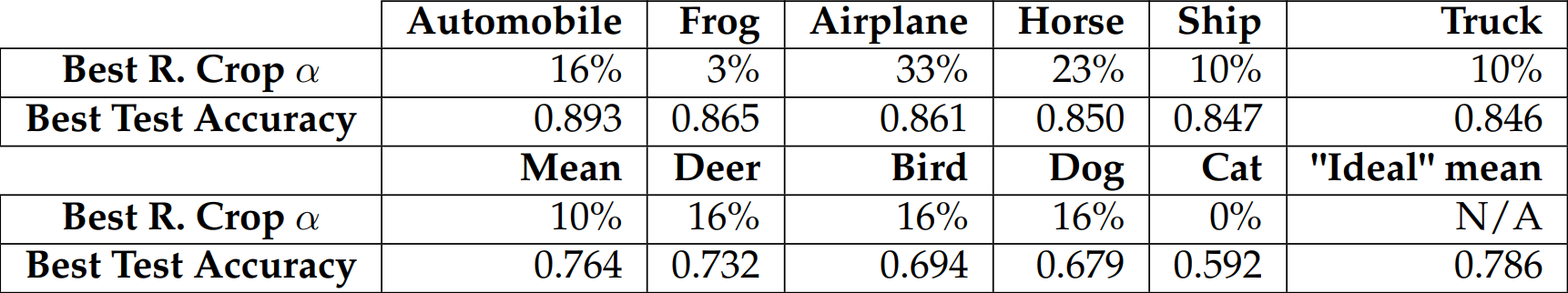}
\end{figure} 
\pagebreak
\appendixsection{Per-class and overall test set performances samples for the Fashion-MNIST + EfficientNetV2S + Random Cropping + Random Horizontal Flip experiment} \label{apx:ninth}
\begin{figure}[H]
\centering
\includegraphics[width=1\textwidth]{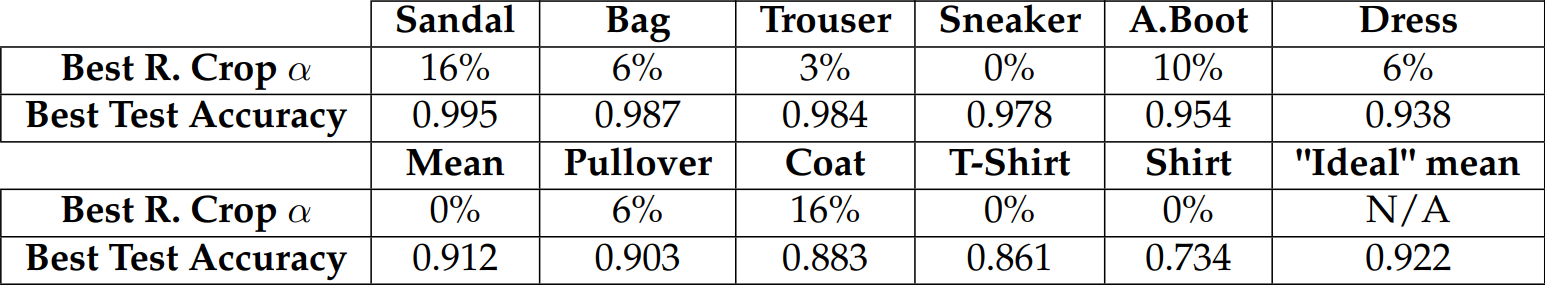}
\end{figure} 
\appendixsection{Per-class and overall test set performances samples for the Fashion-MNIST + ResNet50 + Random Cropping experiment} \label{apx:seventh}

\begin{figure}[H]
\centering
\includegraphics[width=1\textwidth]{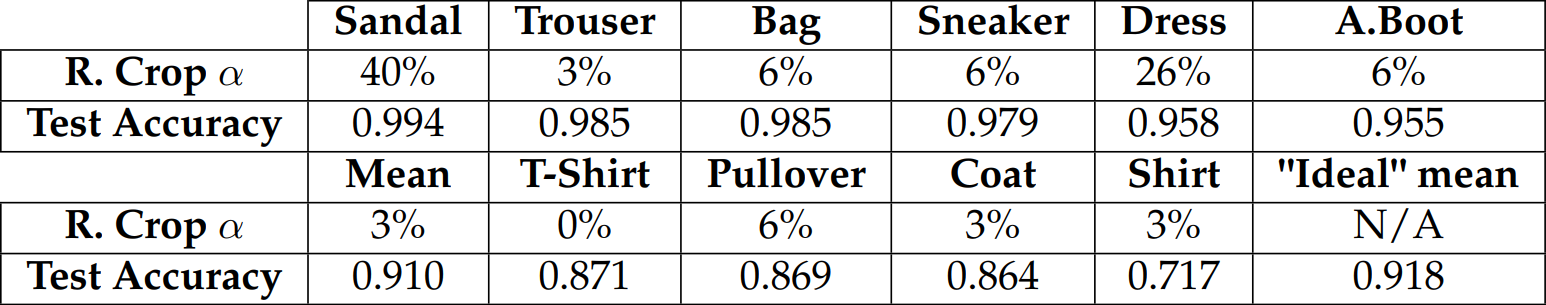}
\end{figure} 

\appendixsection{Per-class and overall test set performances samples for the CIFAR-10 + ResNet50 + Random Cropping experiment} \label{apx:eleventh}

\begin{figure}[H]
\centering
\includegraphics[width=1\textwidth]{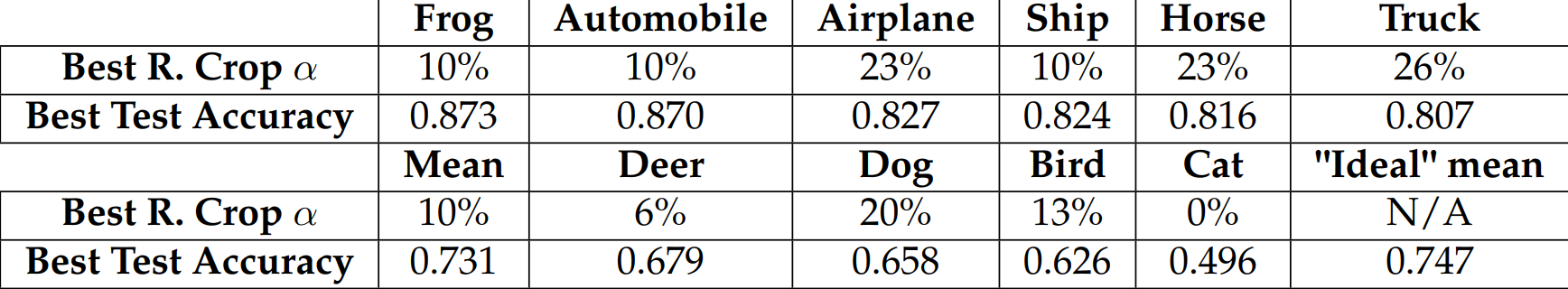}
\end{figure} 

\appendixsection{Per-class and overall test set performances samples for the Fashion-MNIST + SWIN Transformer + Random Cropping + Random Horizontal Flip experiment} \label{apx:tenth}

\begin{figure}[H]
\centering
\includegraphics[width=1\textwidth]{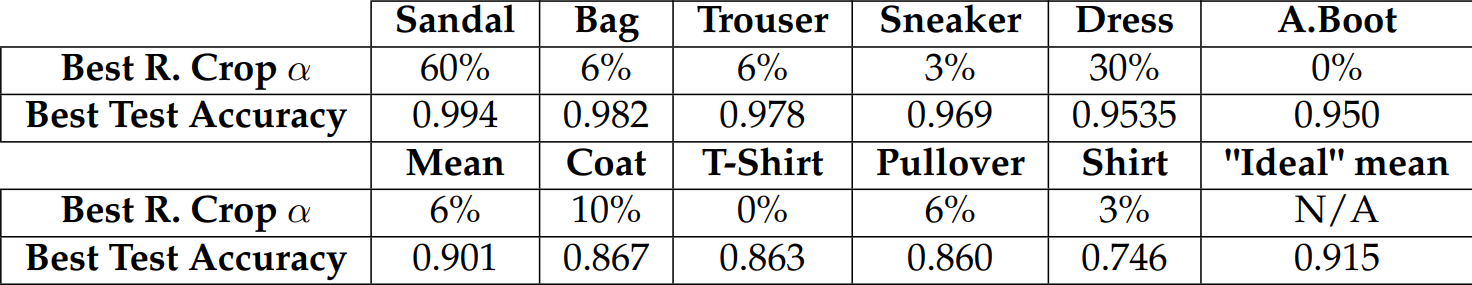}
\end{figure}

%\appendixsection{Link to a Google Drive folder containing all experiment results: model weights, class accuracies, tuning records, training histories, etc.} 
%\label{apx:seventh}
%https://drive.google.com/drive/folders/19TG4xjKkmNhodPSVQ1RopqAwQy5H0U3V
%RIP Experiment Results folder, you were beautiful but extremely expensive to keep around - if need, in the future we can spend some money on compute (or use my M3 work laptop) to rerun these experiments if desired.

\end{document}